\documentclass[12pt,preprint]{elsarticle} 

\usepackage{amssymb,amsthm,amsfonts,latexsym,cancel}
\usepackage[intlimits]{amsmath}
\usepackage{amsfonts}
\usepackage{latexsym}
\usepackage{titlesec}
\usepackage[english]{babel}
\usepackage{epstopdf}
\usepackage{epsfig}
\usepackage{caption}
\usepackage{hyperref}
\usepackage{longtable,multirow,booktabs}
\usepackage{stmaryrd}
\SetSymbolFont{stmry}{bold}{U}{stmry}{m}{n}
\allowdisplaybreaks
\usepackage{graphicx} 
\usepackage{colortbl}  
\usepackage{pifont}    
\usepackage[table]{xcolor}
\usepackage{rotating}
\usepackage{lscape}   

\usepackage{dsfont}
\usepackage{mathrsfs}
\usepackage{eucal} 

\usepackage{geometry}
\usepackage{todonotes} 
\usepackage{fullpage}
\usepackage{dirtree} 

\definecolor{olcgremove}{RGB}{200,30,30}
\definecolor{olcgadd}{RGB}{30,130,30}

\journal{Elsevier}
\begin{document}
\begin{frontmatter}

\title{Learning Unsteady Aneurysm Hemodynamics with Physics-Informed DeepONets}

\author[label1,label2]{Oscar L. Cruz-Gonz\'{a}lez} 
\author[label1]{Val\'{e}rie Deplano\corref{cor1}}\ead{valerie.deplano@univ-amu.fr}
\author[label2]{Badih Ghattas}

\address[label1]{Aix Marseille Univ, CNRS, Centrale Marseille, IRPHE UMR 7342, Marseille, France}
\address[label2]{Aix Marseille Univ, CNRS, AMSE UMR 7316, Marseille, France}\cortext[cor1]{Corresponding author}


\begin{abstract}

Clinically actionable, patient-specific hemodynamic assessment, specifically wall shear stress, vortex structure and pressure distributions, is critical for determining risky or unfavorable evolution in Abdominal Aortic Aneurysms (AAA). While Physics-Informed Deep Operator Networks (PI-DeepONets) show promising results in complementing established tools such as Computational Fluid Dynamics (CFD), a persistent architectural challenge remains for complex 3D flows. 
In this direction, we propose a Modified Multi-Input Multi-Output PI-DeepONets (M$^{3}$PI-DeepONet) designed for predicting unsteady flows in an idealized AAA geometry.  Central to our model is the Aggregated Injection strategy, where latent representations from multiple input branches are fused prior to trunk injection, allowing the coordinate basis to adapt to multiple physical constraints. To the best of our knowledge, this is the first architecture to combine the layer-wise gating mechanism with a multi-branch operator-network topology, yielding an input-adaptive trunk basis. Additionally, we integrate the 3D Navier-Stokes equations as governing physical laws, so the model is trained based on physics-informed residuals, initial and boundary conditions, and only 0.3\% of the labeled internal data together with the selected branch-conditioning signals.
The M$^{3}$PI-DeepONet simultaneously predicts unsteady 3D flow velocity and pressure fields
with an average relative $L_2$ velocity error below 4\% and pressure error around 5\% while achieving a conservative retained-cycle inference speedup of approximately $36\times$ compared to reference CFD simulations once the branch inputs used for conditioning are available.
This work advances the application of deep learning in cardiovascular disease modeling, marking step toward real-time, non-invasive clinical diagnostics.

\end{abstract}


\end{frontmatter}

\newpage
\section{Introduction}\label{sect:intro}


Abdominal Aortic Aneurysm (AAA) unfavorable evolution is a catastrophic event with high mortality rates when rupture occurs. Current clinical decision-making for surgical intervention relies on maximum aneurysm diameter or its rate of growth, geometric criterions that are not enough discriminants. Hemodynamic markers, such as Wall Shear Stress (WSS), pressure, and vortex structure dynamics also drive AAA evolution \cite{boyd2016low,singh_association_2023}. Integrating these biomarkers into clinical workflows is essential for personalized risk detection \cite{qiu2022association, susar2025characterization}. However, obtaining high-fidelity, time-resolved hemodynamic data within the strict limitations of the clinical setting remains a significant downside, requiring a shift toward predictive tools.


The central challenge lies in resolving the interplay between blood flow and arterial mechanics without incurring prohibitive costs. While 4D flow MRI enables non-invasive measurement of cardiovascular flow, its spatial and temporal resolution is limited, especially near vessel walls, resulting in uncertainty in wall location and incorrect estimation of wall shear stress due to insufficient velocity gradient resolution \cite{Jarral2020}. In contrast, computational approaches such as Computational Fluid Dynamics (CFD) provide high-resolution and physiologically consistent representations of hemodynamics and arterial deformation. However, these methods are computationally expensive, with patient-specific simulations often requiring hours to days or longer to be completed, limiting their routine clinical applicability \cite{song2023systematic}.


Recent progress in Scientific Machine Learning (SciML) have mostly focused on making individual models more reliable and combining them with traditional computational methods. In the domain of Physics-Informed Neural Networks (PINNs), researchers have introduced novel architectures to capture complex dynamics; for instance, \cite{wang2025physicsinformedfourierbasisneural} developed Physics-informed Fourier Basis Networks (FBNN) for periodic modeling, while \cite{LIU2025ICPINN} proposed Integral Conservation PINNs (ICPINN) to enforce global conservation laws in 3D blood flow simulations. These architectural advances are complemented by theoretical improvements that address training pathologies, such as second-order optimization for gradient conflict mitigation \cite{wang2025gradientalignmentphysicsinformedneural} and rigorous analyses linking ill-conditioning to PDE Jacobians \cite{cao2024analysissolutionillconditioningphysicsinformed}. Furthermore, the field is increasingly adopting hybrid models that combine neural networks with traditional solvers to handle multi-scale physics. Examples include the Multi-scale Neural Computing (MSNC) framework \cite{suo_novel_2024}, which decomposes solutions between neural networks and finite difference methods, and Physics-Informed Graph Neural Networks (GNNs) that integrate geometric deep learning with physical laws \cite{ZHANG2025109462}. Other approaches, such as the neural differentiable modeling for turbulence \cite{FAN2025117478} and purely constraint-based Fourier Neural Operators like LESnets \cite{ZHAO2025114125}, further demonstrate the push towards more robust, physically grounded solvers.

Despite successes in single-instance solving, there is a significant paradigm shift toward learning generalized solution operators to enhance both computational efficiency and generalizability. A prominent example of this shift are Deep Operator Networks (DeepONets), which are architected to learn continuous solution operators, mappings between infinite-dimensional function spaces, rather than approximating a single function mapping inputs to outputs \cite{LuLu2021DeepONet}. This operator-centric framework is particularly advantageous for parametric Partial Differential Equations (PDEs). While standard Physics-Informed Neural Networks (PINNs) typically require iterative retraining for any variation in source terms, unseen geometries, or boundary conditions, DeepONets provide rapid predictions across these variations. Building upon this foundation, the Physics-Informed Deep Operator Network (PI-DeepONet) represents a transformative advancement, eliminating the total dependency on labeled data by embedding physical laws directly into the loss function, analogous to PINNs \cite{Wang2021DeepOnets, Goswami2023}. This approach synthesizes the best of both paradigms, achieving broad generalizability and strict physical consistency. Recent work has focused on improving the expressiveness and training speed of these operators. \cite{Mandl_2025} introduced Separable DeepONets, a framework that reduces training time by two orders of magnitude while maintaining accuracy, while \cite{wu2025pockanphysicsinformeddeepoperator} enhanced function approximation capabilities by embedding Rational Kolmogorov-Arnold Networks within the PO-CKAN architecture. To bridge the gap with classical methods, researchers have also developed hybrid operator architectures, such as the FEM-DeepONet \cite{kara2025physicsinformeddeeponetcoupledfem} for adaptive mesh coupling and the Physics-Informed Fourier-DeepONet \cite{LIU2026106026}, which synergizes the strengths of DeepONets and Fourier Neural Operators (FNOs). Further innovations in training paradigms, such as ``one-shot" learning methods that minimize reliance on large simulation datasets \cite{jiao_one-shot_2025} and long-time integration strategies \cite{Wang2023Long}, underscore the rapidly expanding potential of operator learning. These developments highlight the DeepONet architecture as a particularly promising candidate for solving parametric PDEs efficiently.

A critical consideration in extending these methods to unsteady fluid dynamics is the architectural treatment of the temporal domain within operator learning frameworks. The most direct paradigm, often termed Space-Time DeepONet, treats time as an intrinsic coordinate within the trunk network, enabling the continuous learning of the solution manifold $u(x,t)$ from initial data in a single forward pass \cite{LuLu2021DeepONet}. This approach has been scaled to climate-level domains by Latent DeepONets \cite{Kontolati2024}, which integrated spatio-temporal features into compact representations. In addition, alternative strategies have emerged to address specific temporal complexities, such as, Sequential DeepONets that replace the branch network with recurrent units like GRUs to capture path-dependent histories \cite{He2024A,Junyan2024}, while Auto-DeepONet frameworks adopt an autoregressive looping strategy to maintain stability over long horizons \cite{luo2024cfdbench}. Further distinguishing temporal features, architectures like the Temporal Neural Operator \cite{diab2025temporal} and 2U-DeepONet \cite{Diab2024} utilize dedicated temporal branches to enhance long-range extrapolation, whereas GraphDeepONet \cite{cho2024GNN} and recent Transformer-based operators \cite{shih2024transformers} redefine message passing to handle time on irregular meshes.



In fluid dynamics, pressure and velocity components exhibit their own physical characteristics and boundary behaviors.  When modeling these coupled fields via DeepONets, a single static trunk network may be restrictive because it represents all outputs using the same coordinate-dependent basis functions, independently of the branch input conditions. Allowing these basis functions to adapt to the branch inputs is therefore a natural way to improve the representation of coupled flow fields. Within the DeepONet literature, two relevant lines of work are especially important for this objective:

\begin{itemize}
  
\item [(I)] Architecture Improvements (Single-Input): The authors in  \cite{Wang2022Improved} explicitly noted that in a conventional DeepONet, inputs are only merged at the very last layer, which can lead to inefficient information fusion. So then, they introduced the Modified PI-DeepONet, demonstrating that layer-wise gating mechanism mitigates gradient issues and signal vanishing. This creates an Input-Adaptive Basis, where the trunk's basis functions dynamically morph based on the branch input function. However, this specific improvement was designed for a single-branch input function and a single-trunk network, respectively. 

\item [(II)] Multi-Input Strategies (Static Basis): The authors in \cite{Jin2022MultipleInput} successfully expanded the operator framework to accommodate multiple branch inputs (e.g., velocity and pressure) via the Multiple-Input Operator Network (MIONet). MIONet was designed to overcome the limitation of operators defined on a single Banach space, allowing for multiple input functions. However, in the MIONet architecture, the outputs of the $n$ branch nets and the trunk net are completely disconnected until they are merged at the very end via a Hadamard product and summation. Because the trunk net never ``sees" the input functions during its forward pass, its basis functions are rigidly fixed and static.  Consequently, the trunk relies on a Static Basis regardless of the branch inputs.
\end{itemize}

In this work, we introduce a Modified Multi-Input Multi-Output PI-DeepONet (hereafter referred to as M$^{3}$PI-DeepONet) to predict unsteady flow within an idealized 3D Abdominal Aortic Aneurysm (AAA) geometry. In this way, we seek to bridge these two methodologies, extending the layer-wise gating mechanism proposed by \cite{Wang2022Improved} to the multi-branch MIONet architecture \cite{Jin2022MultipleInput}. Direct application of this modified- multi-input setting introduces a dimensional challenge regarding which branch features should guide the trunk's adaptation. To resolve this practically, our primary methodological contribution is an Aggregated Injection strategy. By calculating a weighted fusion of the latent representations from multiple active input branches (e.g., inlet velocity and outlet pressure) prior to trunk injection, we provide the trunk with a unified physical context. This integration shifts the architecture away from relying on a static coordinate basis. In the particular case of the AAA geometry, this allows the coordinate basis to adapt to the joint information carried by velocity and pressure channels. This formulation, combined with the regularizing effect of physics-informed Navier-Stokes residuals, allows the network to capture more complex physical couplings and generate predictions from sparse data. To the best of our knowledge, this work represents the first peer-reviewed application explicitly combining the layer-wise gated mechanism with a multi-branch MIONet architecture.

Additionally, this research aims to assess the accuracy of the proposed framework in predicting the unsteady 3D velocity and pressure fields and to evaluate its computational efficiency relative to classical Computational Fluid Dynamics (CFD) approaches. This work constitutes a natural extension of our recent study \cite{cruzgonzalez2026}, which validated baseline operator learning frameworks on the same idealized 3D AAA geometry under steady flow conditions. Preliminary results on the extension to unsteady flow using PINNs and a standard MI-MO-PI-DeepONet architecture, without the Aggregated Injection strategy introduced here, were presented in \cite{cruzgonza_2025mbj}. The present work advances both the physical fidelity of the problem, by considering the unsteady Navier-Stokes equations, and the architectural framework, through the novel M$^{3}$PI-DeepONet. We acknowledge that this idealized geometry does not yet capture the full complexity of in vivo AAA hemodynamics, including fluid-structure interaction (FSI) effects (see \cite{Zhu2022}). Nonetheless, establishing a time-dependent operator learning framework on a well-characterized benchmark is a critical prerequisite for the ultimate goal of enabling fast, patient-specific hemodynamic modeling.

The methodological contribution of this work is twofold. First, previous applications of Physics-Informed DeepONets in cardiovascular biomechanics have predominantly focused on one-dimensional or reduced-order settings, such as arterial blood-flow/blood-pressure modeling and compliant stenosed-vessel surrogates (see \cite{Li2024A,Velikorodny2025DeepOperator}). Recent progress has reached 3D cardiovascular geometries \cite{macraild_accelerated_2024,rabeh20253d,cruzgonzalez2026}; however, applications to unsteady three-dimensional hemodynamics in pathological vascular geometries remain relatively limited in the operator-learning literature. This study therefore extends the operator-learning framework to an unsteady 3D flow within an idealized AAA geometry. Second, we introduce an Aggregated Injection strategy based on a learned weighted fusion of the latent representations from all active branches, injected layer-wise into the trunk network. This converts the trunk's coordinate basis from static to input-adaptive, thereby enabling a multi-branch PI-DeepONet to fuse heterogeneous conditioning inputs and overcoming the optimization issues that degrade naive multi-branch settings.

The manuscript is organized as follows. In Section \ref{sect:methodology}, we formulate the unsteady 3D AAA flow problem, describe the CFD simulations used to generate reference solutions, introduce the proposed M$^{3}$PI-DeepONet and its loss functions, and then detail the dataset construction process. In Section \ref{sect:results}, we report the numerical evaluation of the proposed framework, first identifying the most informative branch inputs and isolating the contribution of the Aggregated Injection strategy through an ablation study, then assessing the model's accuracy in inferring complete velocity profile-dependent datasets for new waveforms not seen during training and quantifying its speedup relative to the reference CFD solver. Finally, in Section \ref{sect:conclusions}, we discuss the implications of our findings, address the limitations of the current study, and outline potential directions for future research.

\section{Methodology}\label{sect:methodology}

\subsection{Problem Formulation and CFD Ground Truth}\label{sect:methodology:2_1}

\subsubsection{Unsteady AAA flow problem formulation}\label{sect:methodology:formulation}

\begin{figure}[htbp]
	\centering	
	\resizebox{\columnwidth}{!}{\includegraphics[scale=0.5]{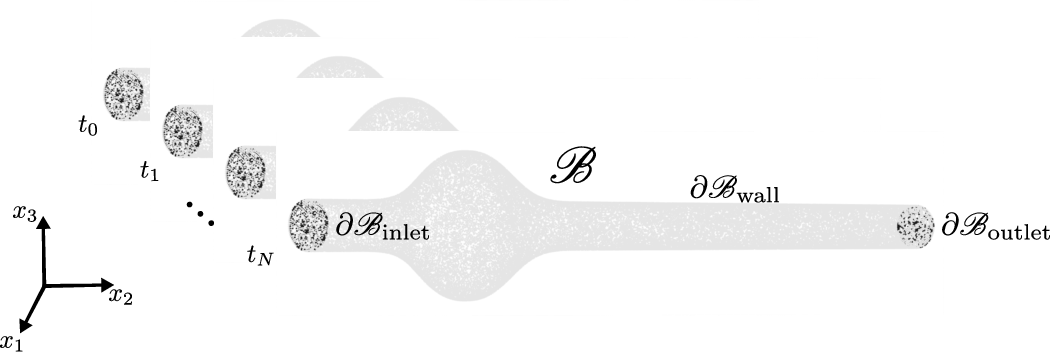}}
	\caption{Abdominal Aortic Aneurysm Idealized Geometry.}
	\label{fig:formulation:aaa}
\end{figure}

In this study, we model the unsteady flow within an idealized three-dimensional Abdominal Aortic Aneurysm (AAA) geometry (see Fig.~\ref{fig:formulation:aaa}). The computational fluid domain, denoted by $\mathscr{B}\subset \mathbb{R}^3$, geometrically represents the lumen of the arterial vessel. This domain is spatially bounded by three distinct surfaces: the proximal inlet $\partial \mathscr{B}_{\text{inlet}}$ where blood enters, the distal outlet $\partial \mathscr{B}_{\text{outlet}}$, and the lateral rigid arterial wall $\partial \mathscr{B}_{\text{wall}}$.

The blood flow dynamics are governed by the 3D Navier-Stokes equations for an incompressible Newtonian fluid. This assumption is considered valid for blood flow in large arteries where the shear rates are sufficiently high. Thus, for spatial coordinates $\boldsymbol{x} \in \mathscr{B}$ and time $t \in (0, \mathscr{T})$, the conservation equations for momentum and mass governing the velocity and pressure fields are formulated as:
\begin{subequations}
	\begin{align}
        & \rho_f\left(\frac{\partial \boldsymbol{v}(\boldsymbol{x},t)}{\partial t}+\left(\boldsymbol{v}(\boldsymbol{x},t) \cdot \nabla\right) \boldsymbol{v}(\boldsymbol{x},t )\right) = - \nabla p(\boldsymbol{x},t ) + \mu_f \nabla^2\boldsymbol{v}(\boldsymbol{x},t ),  \label{momentum3D}\\
        & \nabla\cdot\boldsymbol{v}(\boldsymbol{x},t) = 0,   \label{mass3D}
    \end{align}
\end{subequations}
where $\boldsymbol{v}$ denotes the velocity vector [m/s], $p$ is the scalar pressure [Pa], $\rho_f$ is the constant fluid density [kg/m$^3$], and $\mu_f$ is the dynamic viscosity of the blood [kg/(ms)]. As observed, there are four independent variables which are the spatial coordinates $x_1$, $x_2$, and $x_3$ and the time domain $t$, and four dependent variables, the three components of the velocity $\boldsymbol{v} := (v_1(x_1,x_2,x_3,t),v_2(x_1,x_2,x_3,t),v_3(x_1,x_2,x_3,t))$ and the pressure $p := p(x_1,x_2,x_3,t)$.

To ensure a well-posed problem, the system is closed with specific boundary conditions. First, at the inlet region ($\boldsymbol{x} \in \partial \mathscr{B}_{\text{inlet}}$), a time-dependent velocity profile is prescribed perpendicular to the inlet plane to drive the flow, denoted as $\boldsymbol{v}(\boldsymbol{x}, t) = \boldsymbol{v}^{\{\text{inlet}\}}(\boldsymbol{x}, t)$ for $t \in (0, \mathscr{T})$, where
\begin{align}
 v_1^{\{\text{inlet}\}} = 0, \quad 
 v_2^{\{\text{inlet}\}} = u\!\left(\sqrt{x_1^2 + x_3^2},\, t\right), \quad 
 v_3^{\{\text{inlet}\}} = 0.  \label{womersley_profile}
\end{align}

Here, $r:=\sqrt{x_1^2 + x_3^2}$ is the radial distance from the center of the inlet. To mimic the pulsatile nature of arterial blood flow, $u$ is derived from real patient flow rate $Q(t)$ \cite{Baudouard2024Aortic, 
Baudouard2025Assessing} using the Womersley solution, assuming that the flow is fully developed.

Second, as we assume rigid wall boundary conditions, a no-slip condition is applied at the arterial walls. 
\begin{align}
 \boldsymbol{v}(\boldsymbol{x}, t) = \boldsymbol{0}, \quad t \in (0, \mathscr{T}),\; \boldsymbol{x} \in \partial \mathscr{B}_{\text{wall}}.\label{non_slip}
\end{align}

Additionally, a standard outflow condition is imposed at the distal outlet, assuming fully developed flow with zero normal gradient of the velocity field,
\begin{align}
    \frac{\partial \boldsymbol{v}(\boldsymbol{x}, t)}{\partial \mathbf{n}} = \mathbf{0}, \quad t \in (0, \mathscr{T}), \; \boldsymbol{x} \in \partial \mathscr{B}_{\text{outlet}}, \label{outflow}
\end{align}
where $\boldsymbol{n}$ is the outward unit normal.

Because the velocity profile at the inlet is periodic, say with a physical period $\mathscr{T}$, the hemodynamic regime of interest is the periodic phase state in which the
flow reproduces itself from one cardiac cycle to the next. This target regime satisfies,
\begin{align}
\boldsymbol{v}(\boldsymbol{x}, t) = \boldsymbol{v}(\boldsymbol{x}, t + \mathscr{T}),
\quad p(\boldsymbol{x}, t) = p(\boldsymbol{x}, t + \mathscr{T}),
\quad \boldsymbol{x} \in \mathscr{B}, \; t \geq 0.
\label{eq:periodicity}
\end{align}

\subsubsection{Ground-truth CFD generation}\label{sect:methodology:GT_generation}
The numerical simulations were performed using the Fluent fluid solver (finite volume method) within ANSYS Workbench R2 2023 (ANSYS Inc, Canonsburg, USA). Blood was modeled as a Newtonian fluid with the following physiological properties: a density $\rho_f = 1060\;[\text{kg/m}^3]$ and a dynamic viscosity $\mu_f = 0.00399\;[\text{kg/(m\,s)}]$.

The idealized AAA geometry was constructed to represent a fusiform aneurysm, characterized by the following dimensions: an inlet radius $R_{\text{inlet}} = 0.010065\;[\text{m}]$, an aneurysm radius $R_{\text{AAA}} = 0.03\;[\text{m}]$ (corresponding to a $3\times$ dilatation ratio), and a total vessel length $L = 0.26\;[\text{m}]$. The specimen is oriented so that the center of the inlet region coincides with the origin of the Cartesian coordinate system, with the main flow direction aligned along the $x_2$-axis. A straight tube from the distal aneurysm neck to the domain outlet with a length of $L_{\mathrm{out}} =0.16 [m]$, corresponding to $16R_{inlet}$ was added. This extension is sufficiently long for the flow to be fully developed before the zero-normal-gradient outflow condition is imposed. To resolve spatial gradients accurately, seven near-wall mesh elements, representing 0.91 [mm] thickness, were placed in the wall-normal direction to capture the boundary layer, whose thickness was estimated as $\delta = R_{\text{inlet}}/\alpha \approx 0.769$~[mm] according to Fung's definition \cite{Fung1997}. This meshed near-wall thickness enables the velocity gradients at the rigid wall to be evaluated with sufficient accuracy. The whole domain was discretized into a mesh of $1{,}186{,}076$ elements.

\begin{figure}[htbp]
	\centering	
	\resizebox{15.2cm}{!}{\includegraphics[scale=0.5]{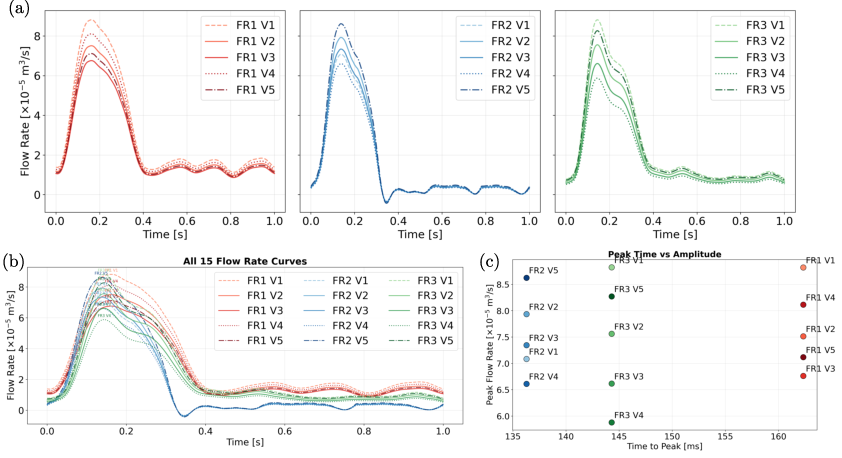}}
	\caption{Flow rate waveforms $Q(t)$ derived from clinical data. (a) Waveforms grouped by patient source: FR1, FR2, and FR3 (five waveforms per group). (b) All 15 flow rate curves overlaid on a single plot, illustrating the full range of temporal and amplitude variability across patient groups. (c) Scatter plot of peak time versus peak flow rate amplitude for each waveform.}
	\label{fig:patients_group}
\end{figure}

To ensure that the model is robust against physiological variability and to enrich the training distribution, we have built a database of 15 distinct flow rate waveforms $Q(t)$. These were derived from clinical data belonging to three patient groups, labeled FR1, FR2, and FR3, with five waveforms per group (see Figure \ref{fig:patients_group}). All three groups share a common temporal profile within each group but differ in peak amplitude. When combined, the 15 flow rate curves illustrate a full range of temporal and amplitude variability across the patient groups. These 15 waveforms served as the basis for constructing the time-varying inlet velocity profiles $\boldsymbol{v}^{\{\text{inlet}\}}(\boldsymbol{x}, t)$.

To obtain a periodic solution, four consecutive cardiac cycles (each of physical period $\mathscr{T}= 1\;\text{s}$), were performed. Each cardiac cycle was discretized into 1000 computational time steps per cycle ($\Delta t = 0.001\;\text{s}$), sufficient to the solution convergence. The solver is initialized from a quiescent state ($\boldsymbol{v}\equiv\boldsymbol{0}$) at the beginning of the first cycle except at the inlet, where the velocity profile at t = 0 s is specified. The fourth cycle, which satisfies the periodicity condition Eq. \eqref{eq:periodicity} within the solver's convergence tolerance, is preserved for the construction of the full dataset used to train and test the operator. The convergence criterion required a root mean square residual error below $10^{-4}$ at every time step. From the fourth cycle of each simulation, a subset of 101 uniformly spaced time steps were extracted for the final dataset ($t_0$ through $t_{100}$, with $t_0 \equiv t_{100}$).  Throughout the remainder of the paper, the initial condition at $t=0$ refer to the velocity and pressure snapshot at the start of this converged fourth cycle, and is subsequently used, in whole or in part, by the operator learning framework (Section~\ref{sect:methodology:network_arch}).

Additionally, the computational cost of the retained fourth cycle was approximately 12 hours, a figure largely dominated by the writing of its 101 saved snapshots to disk rather than by the time integration itself. The three first cycles, whose snapshots are not stored, are faster. This retained-cycle cost is used explicitly in the computational-efficiency analysis.

\subsection{The M$^{3}$PI-DeepONet Formulation}\label{sect:methodology:network_arch}

\subsubsection{Operator Learning Foundation}\label{sect:methodology:network_arch:deeponet_foundation}

Operator Learning addresses a fundamental limitation of Physics-Informed Neural Networks, i.e., the need to retrain the model whenever problem parameters change. Instead of learning a single PDE solution, it learns the mapping between infinite-dimensional function spaces, enabling instant predictions for entire families of PDEs.

Consider a parametric PDE system defined on an open spatial domain $\mathscr{B} \subset \mathbb{R}^d$ with closure $\overline{\mathscr{B}}$, boundary $\partial\mathscr{B}$ and temporal domain $[0, \mathscr{T}]$:
\begin{subequations}
\begin{align}
	\mathfrak{N}[u^{(i)}(\boldsymbol{x}, t);\, f^{(i)}(\boldsymbol{x})] &= 0, \quad \boldsymbol{x} \in \mathscr{B},\; t \in (0, \mathscr{T}], \label{eq:parametric_pde_a}\\
	\mathfrak{B}[u^{(i)}(\boldsymbol{x}, t)] &= 0, \quad \boldsymbol{x} \in \partial\mathscr{B},\; t \in (0, \mathscr{T}], \label{eq:parametric_pde_b}\\
	u^{(i)}(\boldsymbol{x}, 0) &= u_0(\boldsymbol{x}), \quad \boldsymbol{x} \in \overline{\mathscr{B}}, \label{eq:parametric_pde_c}
\end{align}
\end{subequations}
 where $\{f^{(i)}\}_{i=1}^{N}$ is a family of $N$ input parameters and $u^{(i)}$ the associated PDE solution function. The operator $\mathfrak{N}$ represents the general PDE residual in the interior of the domain, and $\mathfrak{B}$ represents the boundary residual operator on $\partial\mathscr{B}$. The objective is to learn the operator $\mathfrak{G}\colon \mathscr{X} \to \mathscr{Y}$ between infinite-dimensional Banach spaces such that $u^{(i)}(\boldsymbol{x}, t) = \mathfrak{G}(f^{(i)})(\boldsymbol{x}, t)$.

Lu et al. \cite{LuLu2021DeepONet} proposed an unstacked deep learning architecture called Deep Operator Network (DeepONet) to approximate $\mathfrak{G}$ based on the Universal Approximation Theorem for Operators~\cite{Chen1995}. The architecture employs two sub-networks: a \textit{Branch net} that encodes the input function $f^{(i)}$ (evaluated at $m$ fixed sensor points $\{\tilde{\boldsymbol{x}}_k\}_{k=1}^m$) into a latent embedding $[\beta_1, \ldots, \beta_q]^T \in \mathbb{R}^q$, and a \textit{Trunk net} that encodes the query coordinates $(\boldsymbol{x}, t)$ into an embedding $[\tau_1, \ldots, \tau_q]^T \in \mathbb{R}^q$. The operator approximation is given by their dot product:
\begin{align}
	\hat{\mathfrak{G}}_\theta(f^{(i)})(\boldsymbol{x}_j^{(i)}, t_j^{(i)})
	\;:=\;
	\sum_{k=1}^{q}
	\underset{\text{Branch}}{\underbrace{\beta_{k}\!\bigl(f^{(i)}(\tilde{\boldsymbol{x}}_1), \ldots, f^{(i)}(\tilde{\boldsymbol{x}}_m)\bigr)}}
	\;\;
	\underset{\text{Trunk}}{\underbrace{\tau_k(\boldsymbol{x}_j^{(i)}, t_j^{(i)})}},
	\label{eq:deeponet_vanilla}
\end{align}
where $\theta$ denotes all trainable parameters in both branch and trunk networks. The standard DeepONet is purely data-driven, minimizing the mean squared error between predictions and ground truth solution data over $N$ input functions and $P$ query points (see~\cite{LuLu2021DeepONet} for more details). Wang et al. \cite{Wang2021DeepOnets} extended this framework to Physics-Informed DeepONets (PI-DeepONets) by incorporating PDE residuals, boundary conditions, and initial conditions into the loss function, analogous to the PINN paradigm, thereby reducing or eliminating the need for labeled solution data.

\subsubsection{Proposed Architecture}\label{sect:methodology:network_arch:proposed_architecture}

The aim of this work is to learn a nonlinear operator $\mathfrak{G}$ that maps selected input functions to the full spatio-temporal solution of the incompressible Navier-Stokes equations within the AAA idealized geometry. We define $\mathcal{S}$ as the set of candidate scalar input channels. These channels are treated as selectable information sources rather than as a fixed set of problem parameters. That is,
\begin{align}
\mathcal{S} \;=\; \bigl\{\, v_2^{\{\text{inlet}\}},\;\; p^{\{\text{inlet}\}},\;\; v_2^{\{\text{outlet}\}},\;\; p^{\{\text{outlet}\}},\;\; v_2^{\{\text{ic}\}},\;\; p^{\{\text{ic}\}} \,\bigr\}, \label{eq:candidate_set}
\end{align}
where $\{\text{inlet}\}$ and $\{\text{outlet}\}$ denote quantities sampled on the corresponding boundary regions, while $\{\text{ic}\}$ denotes quantities sampled at the retained $t=0$ phase of the periodic cycle, restricted to the inlet, outlet and wall regions. 

The choice of streamwise velocity $v_2$ and pressure $p$ is a physically motivated modeling assumption rather than a limitation of the proposed formulation. In principle, additional velocity components or other informative quantities could also be included in $\mathcal{S}$. Moreover, the elements of $\mathcal{S}$ should be interpreted as information channels supplied to the operator, and not necessarily as prescribed boundary or initial conditions of the underlying flow problem. For example, $v_2^{{\mathrm{inlet}}}$ may be constructed prospectively from a prescribed or measured flow-rate waveform $Q(t)$ using a Womersley profile. By contrast, quantities such as $p^{{\mathrm{outlet}}}$, $v_2^{{\mathrm{ic}}}$, and $p^{{\mathrm{ic}}}$ are generally not known before solving the flow problem. Among these, the initial-state branch $v_2^{\{\text{ic}\}}$ is only partially unknown a priori: on the inlet boundary it coincides with the prescribed waveform evaluated at $t=0$, and on the rigid wall it vanishes by the no-slip condition, so that only its outlet trace at $t=0$ genuinely requires the converged CFD solution in the present idealized setup.

In the present study, these latter quantities are available because they are extracted from the retained fourth CFD cycle, after the simulation has reached a phase-periodic regime. Therefore, configurations that use such inputs should be interpreted as measurement-conditioned or data-assimilation settings, rather than as surrogates driven solely by prescribed inlet waveform data. In a prospective clinical or CFD-free workflow, these channels would need to be provided by external measurements, a reduced 0D/1D model  (for example, an outlet pressure reconstructed from a Windkessel model driven by the measured outlet flow rate), a reduced-order spin-up procedure, or a learned auxiliary estimator.

So then, given an active subset $\mathcal{S}_{\text{active}} = \{f_1,\ldots,f_n\} \subseteq \mathcal{S}$, we seek to predict the velocity vector field and the pressure scalar field at any spatial location and time instant,
\begin{align}
	\bigl(v^{(i)}_1,\; v^{(i)}_2,\; v^{(i)}_3,\; p^{(i)}\bigr)(\boldsymbol{x}, t) =
	\mathfrak{G} \bigl(f^{(i)}_1, \ldots, f^{(i)}_n\bigr)(\boldsymbol{x}, t).
	\label{eq:operator_goal}
\end{align}

\begin{figure}[htb]
	\centering
	\includegraphics[scale=0.6]{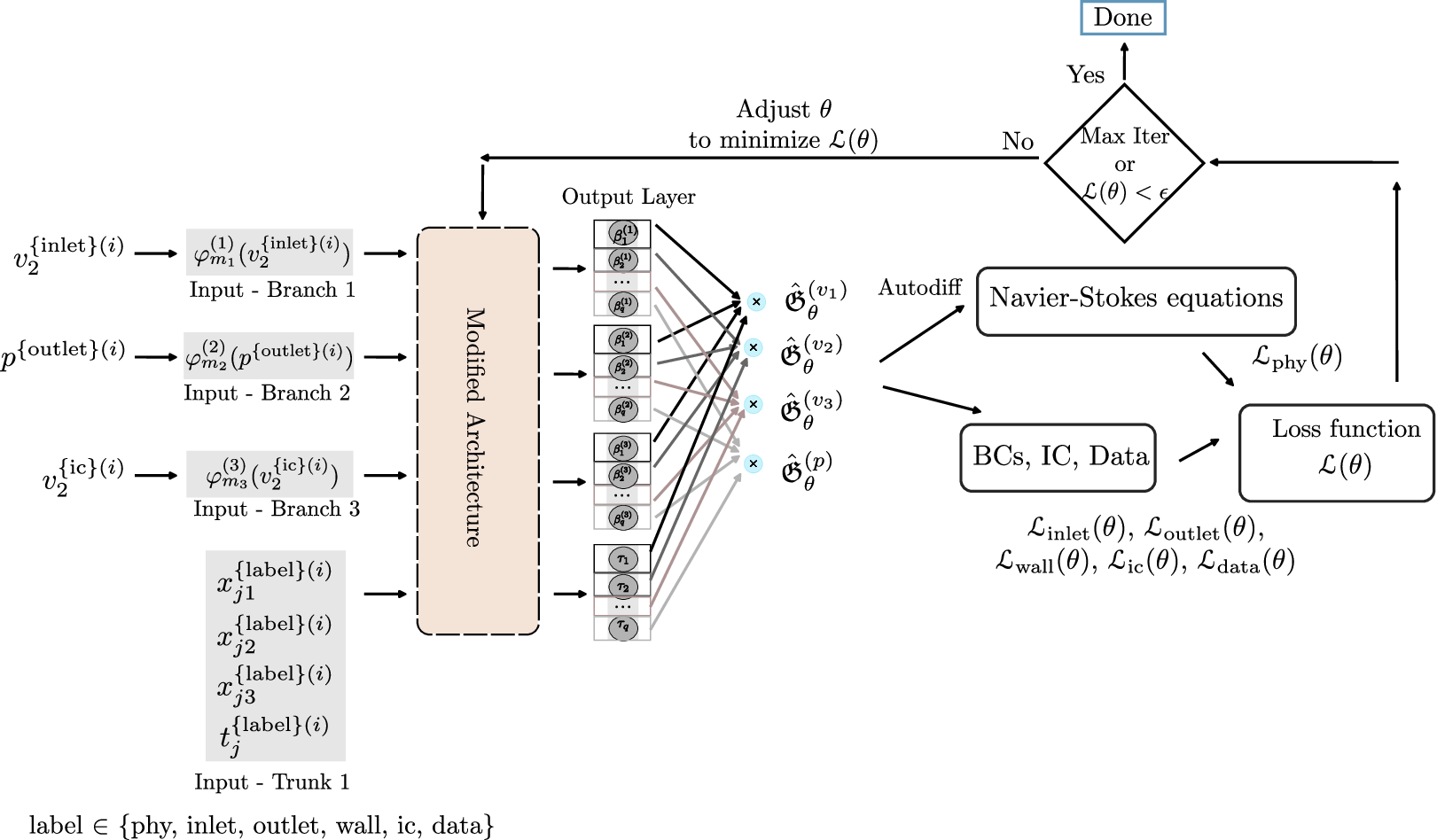}
	\caption{Schematic of the M$^{3}$PI-DeepONet in the physically natural three-branch instantiation ($n = 3$). Branches encode $v_2^{\{\text{inlet}\}(i)}$, $p^{\{\text{outlet}\}(i)}$, and $v_2^{\{\text{ic}\}(i)}$ at their sensor points; the trunk maps $(x_1, x_2, x_3, t)$. Modified architecture with Aggregated Injection (Figure~\ref{fig:adapting:modifiedarch}) condition the coordinate basis on all three inputs jointly. The output layer produces $(\hat{\mathfrak{G}}_\theta^{(v_1)},
 \hat{\mathfrak{G}}_\theta^{(v_2)},
 \hat{\mathfrak{G}}_\theta^{(v_3)},
 \hat{\mathfrak{G}}_\theta^{(p)})$ via output splitting. The training loop minimizes the total loss $\mathcal{L}(\theta)$ defined in Section~\ref{sect:adapting:deeponet:loss}.}
	\label{fig:rotated}
\end{figure}

\begin{figure}[htb]
	\centering	
	\includegraphics[scale=0.5]{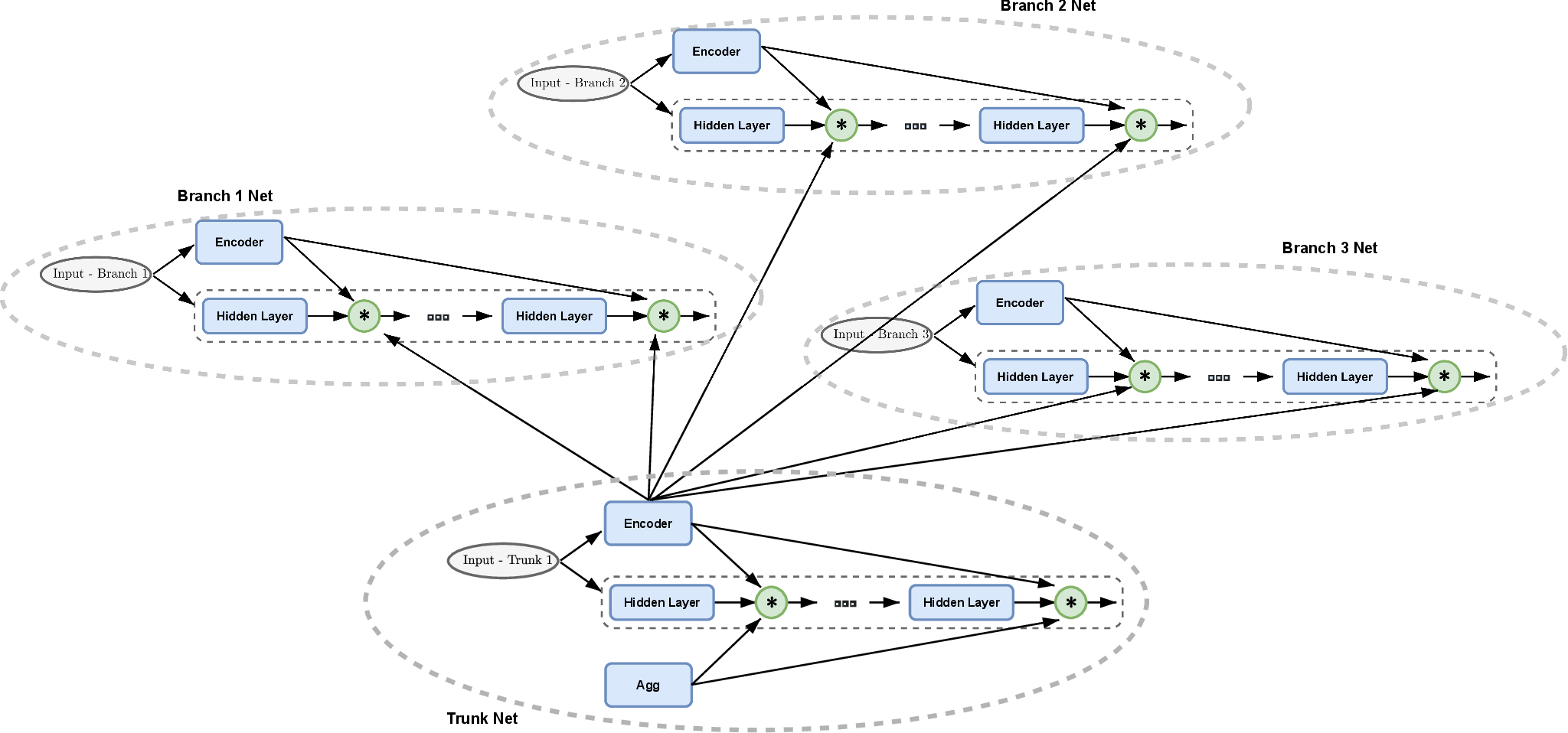}
	\caption{Modified architecture with Aggregated Injection, depicted for the three-branch instantiation ($n = 3$). The forward propagation in conjunction with the gating mechanism is modulated in such a way that each branch net uses its corresponding encoder $\mathbf{U}^{(\xi)}$ together with the trunk encoder $\mathbf{V}_{\text{trunk}}$, while the trunk net uses its corresponding encoder $\mathbf{V}_{\text{trunk}}$ together the aggregated encoder $\mathbf{U}_{\text{agg}}$, endowing the basis functions with their input-adaptive character. The * operation is the one defined in Eqs.\eqref{eq:branch_injection}-\eqref{eq:trunk_injection}.}
	\label{fig:adapting:modifiedarch}
\end{figure}

To approximate this operator, we introduce a M$^{3}$PI-DeepONet (see Figures~\ref{fig:rotated} and~\ref{fig:adapting:modifiedarch}), which bridges two previously independent approaches of operator-learning research: the modified architecture with layer-wise gated mechanism for single-input of Wang et al.~\cite{Wang2022Improved} and the multi-branch MIONet topology of Jin et al.~\cite{Jin2022MultipleInput}, summarized in Table~\ref{tab:architecture_evolution}. Our proposed architecture combines these features through an Aggregated Injection strategy, in which the latent representations of all active input branches are fused via a learnable weighted sum and injected into the trunk network. This yields an input-adaptive coordinate basis, in contrast to the static basis used by standard MIONet. While our previous work \cite{cruzgonzalez2026} applied a multi-input, multi-output PI-DeepONet to the \emph{steady} counterpart of the present problem, and a preliminary unsteady extension without Aggregated Injection was reported in \cite{cruzgonza_2025mbj}, here we fill the gap for the \emph{unsteady} regime.

\begin{table}[htbp]
	\centering
	\caption{Architectural evolution of operator networks toward multi-input adaptive operators. The M$^{3}$PI-DeepONet fills the previously unexplored combination of multi-input encoding with input-adaptive trunk basis functions.}
	\label{tab:architecture_evolution}
	\resizebox{\columnwidth}{!}{
	\begin{tabular}{p{3.8cm} p{3cm} p{7cm} p{3cm}}
		\toprule
		\textbf{Architecture} & \textbf{Input} & \textbf{Trunk Basis Nature} & \textbf{Reference} \\
		\midrule
		Vanilla DeepONet & Single ($f$) & Static (fixed shapes) & Lu et~al.~\cite{LuLu2021DeepONet} \\
		Modified DeepONet & Single ($f$) & \textbf{Adaptive} (depends on $f$) & Wang et~al.~\cite{Wang2022Improved} \\
		Standard MIONet & \textbf{Multi} ($f_1, f_2$) & Static (fixed shapes) & Jin et~al.~\cite{Jin2022MultipleInput} \\
		\textbf{M$^{3}$PI-DeepONet} & \textbf{Multi} ($f_1, \ldots, f_n$, $n$ arbitrary; illustrated below with $n = 3$) & \textbf{Adaptive} (depends on aggregated fusion of $f_1, \ldots, f_n$) & This work \\
		\bottomrule
	\end{tabular}}
\end{table}

Concretely, the M$^{3}$PI-DeepONet integrates three core design elements: (i) $n$ independent input branches, one per active input function in $\mathcal{S}_{\text{active}}$, (ii) a spatio-temporal trunk, and (iii) modified architecture with Aggregated Injection, which condition the coordinate basis on the joint context of all active inputs. All equations below are written for arbitrary $n$. Figures~\ref{fig:rotated} and~\ref{fig:adapting:modifiedarch} use the physically natural three-branch tuple $(v_2^{\{\text{inlet}\}},\, p^{\{\text{outlet}\}},\, v_2^{\{\text{ic}\}})$ only as an illustrative instantiation. It mirrors the natural inlet-flow, outlet-pressure, and initial-state information in this problem, but the architecture itself is not tied to $n=3$.

In what follows, we define the branch networks, trunk network, output splitting, and the Aggregated Injection strategy for this general active-subset formulation.

Based on the multi-input operator framework of Jin et al. \cite{Jin2022MultipleInput}, we employ one independent branch network per active input. Each branch net $\xi \in \{1, \ldots, n\}$ receives its respective input function which is discretized at $m_\xi$ fixed sensor points via the encoding $\varphi_{m_\xi}^{(\xi)}$:
\begin{align}
	\varphi_{m_\xi}^{(\xi)}(f_\xi^{(i)}) = \bigl(f_\xi^{(i)}(\tilde{\boldsymbol{x}}_1^{(\xi)}),\; f_\xi^{(i)}(\tilde{\boldsymbol{x}}_2^{(\xi)}),\; \ldots,\; f_\xi^{(i)}(\tilde{\boldsymbol{x}}_{m_\xi}^{(\xi)})\bigr), \quad \xi = 1, \ldots, n.
	\label{eq:sensor_encoding}
\end{align}
and produces a latent embedding $\boldsymbol{\beta}^{(\xi)} \in \mathbb{R}^q$.

On the other hand, the trunk net receives a four-dimensional input $(x_1, x_2, x_3, t) \in \mathbb{R}^4$ and produces the basis embedding $[\tau_1, \ldots, \tau_q]^T \in \mathbb{R}^q$.

The solution of the unsteady Navier-Stokes equations \eqref{momentum3D}-\eqref{mass3D} comprises four scalar fields, that is, the three velocity components $(v_1, v_2, v_3)$ and the pressure $p$. Rather than training four independent networks, we adopt the multiple-output extension of Wang et al. \cite{Wang2023Long}, which partitions the $q$ trunk-branch interaction modes into $\check{n} = 4$ disjoint slices, one per output component. For the $\zeta$-th output ($\zeta = 1, \ldots, 4$), the operator prediction is
\begin{align}
\hat{\mathfrak{G}}_\theta^{(\zeta)}\!\bigl(f_1^{(i)}, \ldots, f_n^{(i)}\bigr)(\boldsymbol{x}_j^{(i)}, t_j^{(i)})
	\;=\;
	\sum_{k=q_{\zeta-1}+1}^{q_\zeta}
	\tau_k(\boldsymbol{x}_j^{(i)}, t_j^{(i)})
	\prod_{\xi=1}^{n} \beta_k^{(\xi)}\!\bigl(\varphi_{m_\xi}^{(\xi)}(f_\xi^{(i)})\bigr),
	\label{eq:output_splitting}
\end{align}
where $0 = q_0 < q_1 < \cdots < q_4 = q$ defines the partition, and $\{f_\xi\}_{\xi=1}^{n}$ denotes the $n$ active branch inputs. 
For notational convenience, we collect the four scalar components predicted by the operator as
$\hat{\mathfrak{G}}_\theta =
[\hat{\mathfrak{G}}_\theta^{(v_1)},\,
 \hat{\mathfrak{G}}_\theta^{(v_2)},\,
 \hat{\mathfrak{G}}_\theta^{(v_3)},\,
 \hat{\mathfrak{G}}_\theta^{(p)}]$,
where the component index $\zeta \in \{1,2,3,4\}$ refers respectively to
$v_1$, $v_2$, $v_3$, and $p$. This formulation ensures that all four fields share the underlying branch and trunk representations while maintaining distinct latent subspaces for each physical quantity.

As stated before, we adopt the modified architecture proposed by Wang et al. \cite{Wang2022Improved}, which excels in its gating mechanism through the network. For a hidden layer $\mathbf{H}^{(l)}$, the forward propagation is modulated by two context encoders $\mathbf{U}$ and $\mathbf{V}$ in the following way:
\begin{subequations}
\begin{align}
	\mathbf{Z}^{(l)} &= \sigma\!\bigl(\mathbf{W}_z^{(l)} \mathbf{H}^{(l)} + \mathbf{b}_z^{(l)}\bigr), \label{eq:gate_z}\\
	\mathbf{H}^{(l+1)} &= \bigl(1 - \mathbf{Z}^{(l)}\bigr) \odot \mathbf{U} \;+\; \mathbf{Z}^{(l)} \odot \mathbf{V}, \label{eq:gate_update}
\end{align}
\end{subequations}
where $\sigma$ is a nonlinear activation function and $\odot$ denotes element-wise multiplication. Our contribution aims to re-purpose and extend this mechanism to the multi-input setting, where it acquires a fundamentally new role.

Extending the gating mechanism from a single branch to multiple branches introduces a dimensional challenge, that is, the trunk network must be conditioned on the joint physics of the problem, but each branch encodes a different input channel drawn from $\mathcal{S}_{\text{active}}$, with its own characteristic scale, domain, and physical role. Injecting any single branch into the trunk would bias the coordinate basis toward that channel, while direct concatenation would make the trunk-conditioning dimension depend on the number and ordering of active branches. Our \textit{Aggregated Injection} strategy resolves this by first fusing the individual branch embeddings into a single context encoder $\mathbf{U}_{\text{agg}}$ via a learnable weighted  as follows:
\begin{align}
	\mathbf{U}_{\text{agg}} = \sum_{\xi=1}^{n} \alpha_\xi  \mathbf{U}^{(\xi)}, \label{eq:aggregated_context}
\end{align}
where $\mathbf{U}^{(\xi)}$ is the corresponding encoder associated with the $\xi$-th branch, and $\boldsymbol{\alpha} = [\alpha_1, \ldots, \alpha_n]$ are unconstrained trainable scalar weights, initialized uniformly and learned directly alongside the network parameters. The injection then operates asymmetrically across the sub-networks. For the $l$-th hidden layer of the $\xi$-th branch, the gate blends the branch-specific embedding with the trunk embedding, preserving each branch's distinct representation:
\begin{align}
	\mathbf{H}_{\text{branch},\xi}^{(l+1)} = \bigl(1 - \mathbf{Z}_{\text{branch},\xi}^{(l)}\bigr) \odot \mathbf{U}^{(\xi)} \;+\; \mathbf{Z}_{\text{branch},\xi}^{(l)} \odot \mathbf{V}_{\text{trunk}}.
	\label{eq:branch_injection}
\end{align}

For the trunk, the gate blends the aggregated branch information with the trunk's own spatial embedding:
\begin{align}
	\mathbf{H}_{\text{trunk}}^{(l+1)} = \bigl(1 - \mathbf{Z}_{\text{trunk}}^{(l)}\bigr) \odot \mathbf{U}_{\text{agg}} \;+\; \mathbf{Z}_{\text{trunk}}^{(l)} \odot \mathbf{V}_{\text{trunk}}.
	\label{eq:trunk_injection}
\end{align}

By injecting $\mathbf{U}_{\text{agg}}$ directly into the trunk's forward pass, the resulting basis functions $\tau_k(\boldsymbol{x}, t;\, f_1, \ldots, f_n)$ become input-adaptive, the coordinate basis depends explicitly on the active branch tuple rather than only on $(\boldsymbol{x},t)$. In the AAA geometry, each branch therefore retains a dedicated latent representation through Eq. \eqref{eq:branch_injection}, while the trunk adapts to their joint context through Eq. \eqref{eq:trunk_injection}. 

\subsubsection{Physics-Informed Loss Function}\label{sect:adapting:deeponet:loss}
The M$^{3}$PI-DeepONet is trained by minimizing a total loss function that enforces the unsteady Navier--Stokes equations, boundary conditions, initial conditions, and available sparse data,

\begin{align}
	\mathcal{L}_{\text{M$^{3}$PI-DeepONet}}(\theta) &= \lambda_{\text{data}}\,\mathcal{L}_{\text{data}}(\theta) + \lambda_{\text{inlet}}\,\mathcal{L}_{\text{inlet}}(\theta) + \lambda_{\text{outlet}}\,\mathcal{L}_{\text{outlet}}(\theta) \nonumber\\
	& \quad + \lambda_{\text{wall}}\,\mathcal{L}_{\text{wall}}(\theta) + \lambda_{\text{ic}}\,\mathcal{L}_{\text{ic}}(\theta) + \lambda_{\text{phy}}\,\mathcal{L}_{\text{phy}}(\theta). \label{eq:loss_pideeponet}
\end{align}

It should be noted that the neural network is trained using the dimensionless approach described in \ref{appendix:training_protocol}. Therefore, the physical residual loss uses the associated dimensionless Navier-Stokes form, derived in \ref{appendix:dimensionless}. All variables in the formulas below are dimensionless, and asterisk notation is omitted for brevity. In addition, for notational compactness, we use $\hat{\mathfrak{G}}_\theta^{(\cdot)}(\cdot)$ as a shorthand for $\hat{\mathfrak{G}}_\theta^{(\cdot)}\!\bigl(f_1^{(i)}, \ldots, f_n^{(i)}\bigr)(\cdot)$, the illustrative three-branch case $\mathcal{S}_{\text{active}}\ = \bigl(v_2^{\{\text{inlet}\}(i)},\, p^{\{\text{outlet}\}(i)},\, v_2^{\{\text{ic}\}(i)}\bigr)$ is considered, and we write $\hat{\mathfrak{G}}_\theta^{(\boldsymbol{v})} = \bigl(\hat{\mathfrak{G}}_\theta^{(v_1)},\, \hat{\mathfrak{G}}_\theta^{(v_2)},\, \hat{\mathfrak{G}}_\theta^{(v_3)}\bigr)$ for the predicted velocity vector.  Let's look at the definitions of the total loss components.

The data loss enforces agreement for velocity and pressure fields at sparse supervised interior measurement points, sampled from the Volume region and restricted to the training partition (see Section~\ref{sect:methodology:dataset_construction} for more details). That is:
\begin{align}
\mathcal{L}_{\text{data}}(\theta) &= \frac{1}{N P_{\text{data}}}\sum_{i=1}^{N}\sum_{j=1}^{P_{\text{data}}} \Bigl( \bigl\| \hat{\mathfrak{G}}^{(\boldsymbol{v})}_\theta(\boldsymbol{x}^{\{\text{data}\}(i)}_{j}, t^{\{\text{data}\}(i)}_{j}) - \boldsymbol{v}(\boldsymbol{x}^{\{\text{data}\}(i)}_{j}, t^{\{\text{data}\}(i)}_{j})\bigr\|^2 \nonumber \\
& \quad + \bigl| \hat{\mathfrak{G}}^{(p)}_\theta(\boldsymbol{x}^{\{\text{data}\}(i)}_{j}, t^{\{\text{data}\}(i)}_{j}) - p(\boldsymbol{x}^{\{\text{data}\}(i)}_{j}, t^{\{\text{data}\}(i)}_{j})\bigr|^2 \Bigr).\label{eq:loss_data}
\end{align}

The \textit{inlet} and \textit{outlet losses} enforces the prescribed velocity field on the inlet and outlet boundaries, respectively:
\begin{align}
\mathcal{L}_{\text{inlet}}(\theta) &= \frac{1}{N P_{\text{inlet}}} \sum_{i=1}^{N}\sum_{k=1}^{P_{\text{inlet}}} \bigl\| \hat{\mathfrak{G}}^{(\boldsymbol{v})}_\theta(\boldsymbol{x}^{\{\text{inlet}\}(i)}_{k}, t^{\{\text{inlet}\}(i)}_{k}) - \boldsymbol{v}(\boldsymbol{x}^{\{\text{inlet}\}(i)}_{k}, t^{\{\text{inlet}\}(i)}_{k})\bigr\|^2,\label{eq:loss_inlet}
\end{align}
\begin{align}
\mathcal{L}_{\text{outlet}}(\theta) &= \frac{1}{N P_{\text{outlet}}}\sum_{i=1}^{N}\sum_{s=1}^{P_{\text{outlet}}} \bigl\| \hat{\mathfrak{G}}^{(\boldsymbol{v})}_\theta(\boldsymbol{x}^{\{\text{outlet}\}(i)}_{s}, t^{\{\text{outlet}\}(i)}_{s}) - \boldsymbol{v}(\boldsymbol{x}^{\{\text{outlet}\}(i)}_{s}, t^{\{\text{outlet}\}(i)}_{s})\bigr\|^2. \label{eq:loss_outlet}
\end{align}

The \textit{wall loss} enforces the no-slip condition on the vessel wall:
\begin{align}
\mathcal{L}_{\text{wall}}(\theta) &= \frac{1}{NP_{\text{wall}}}\sum_{i=1}^{N}\sum_{l=1}^{P_{\text{wall}}} \bigl\| \hat{\mathfrak{G}}^{(\boldsymbol{v})}_\theta(\boldsymbol{x}^{\{\text{wall}\}(i)}_{l}, t^{\{\text{wall}\}(i)}_{l}) - \boldsymbol{0}\bigr\|^2. \label{eq:loss_wall}
\end{align}

The \textit{initial condition loss} ensures that the predicted solution matches the known velocity at $t = 0$ exclusively at the boundary regions (Inlet, Outlet, and Wall):
\begin{align}
\mathcal{L}_{\text{ic}}(\theta) &= \frac{1}{N P_{\text{ic}}}\sum_{i=1}^{N}\sum_{l=1}^{P_{\text{ic}}} \bigl\| \hat{\mathfrak{G}}^{(\boldsymbol{v})}_\theta(\boldsymbol{x}^{\{\text{ic}\}(i)}_{l}, 0) - \boldsymbol{v}_0(\boldsymbol{x}^{\{\text{ic}\}(i)}_{l})\bigr\|^2. \label{eq:loss_ic}
\end{align}

Finally, the \textit{physics loss} enforces the unsteady, incompressible Navier--Stokes equations at collocation points:
\begin{align}
\mathcal{L}_{\text{phy}}(\theta) &= \frac{1}{N P_{\text{phy}}} \sum_{i=1}^{N}\sum_{r=1}^{P_{\text{phy}}} \bigl\| \hat{\boldsymbol{E}}_{\theta}(\boldsymbol{x}_r^{\{\text{phy}\}(i)}, t_r^{\{\text{phy}\}(i)})\bigr\|^2, \label{eq:loss_phy}
\end{align}
where $\hat{\boldsymbol{E}}_\theta = (\hat{E}_{\theta 1},\, \hat{E}_{\theta 2},\, \hat{E}_{\theta 3},\, \hat{E}_{\theta 4})$ denotes the residual. The first three components correspond to the unsteady momentum equations, and the fourth to the continuity equation. That is:
\begin{subequations}
{\small\begin{align}
	\hat{E}_{\theta 1} &= \frac{\alpha^2}{Re} \frac{\partial \hat{\mathfrak{G}}^{(v_1)}_\theta}{\partial t} + \!\left(\hat{\mathfrak{G}}^{(v_1)}_\theta\frac{\partial \hat{\mathfrak{G}}^{(v_1)}_\theta}{\partial x_1} + \hat{\mathfrak{G}}^{(v_2)}_\theta\frac{\partial \hat{\mathfrak{G}}^{(v_1)}_\theta}{\partial x_2}+ \hat{\mathfrak{G}}^{(v_3)}_\theta\frac{\partial \hat{\mathfrak{G}}^{(v_1)}_\theta}{\partial x_3}\right) + \frac{\partial \hat{\mathfrak{G}}^{(p)}_\theta}{\partial x_1} \nonumber\\
	&\quad - \frac{1}{Re}\!\left(\frac{\partial^2 \hat{\mathfrak{G}}^{(v_1)}_\theta}{\partial x_1^2} + \frac{\partial^2 \hat{\mathfrak{G}}^{(v_1)}_\theta}{\partial x_2^2} + \frac{\partial^2 \hat{\mathfrak{G}}^{(v_1)}_\theta}{\partial x_3^2}\right),\label{eq:residual_1}\\[6pt]
	\hat{E}_{\theta 2} &= \frac{\alpha^2}{Re} \frac{\partial \hat{\mathfrak{G}}^{(v_2)}_\theta}{\partial t} + \!\left(\hat{\mathfrak{G}}^{(v_1)}_\theta\frac{\partial \hat{\mathfrak{G}}^{(v_2)}_\theta}{\partial x_1} + \hat{\mathfrak{G}}^{(v_2)}_\theta\frac{\partial \hat{\mathfrak{G}}^{(v_2)}_\theta}{\partial x_2}+ \hat{\mathfrak{G}}^{(v_3)}_\theta\frac{\partial \hat{\mathfrak{G}}^{(v_2)}_\theta}{\partial x_3}\right) + \frac{\partial \hat{\mathfrak{G}}^{(p)}_\theta}{\partial x_2} \nonumber\\
	&\quad - \frac{1}{Re}\!\left(\frac{\partial^2 \hat{\mathfrak{G}}^{(v_2)}_\theta}{\partial x_1^2} + \frac{\partial^2 \hat{\mathfrak{G}}^{(v_2)}_\theta}{\partial x_2^2} + \frac{\partial^2 \hat{\mathfrak{G}}^{(v_2)}_\theta}{\partial x_3^2}\right),\label{eq:residual_2}\\[6pt]
	\hat{E}_{\theta 3} &= \frac{\alpha^2}{Re} \frac{\partial \hat{\mathfrak{G}}^{(v_3)}_\theta}{\partial t} + \!\left(\hat{\mathfrak{G}}^{(v_1)}_\theta\frac{\partial \hat{\mathfrak{G}}^{(v_3)}_\theta}{\partial x_1} + \hat{\mathfrak{G}}^{(v_2)}_\theta\frac{\partial \hat{\mathfrak{G}}^{(v_3)}_\theta}{\partial x_2}+ \hat{\mathfrak{G}}^{(v_3)}_\theta\frac{\partial \hat{\mathfrak{G}}^{(v_3)}_\theta}{\partial x_3}\right) + \frac{\partial \hat{\mathfrak{G}}^{(p)}_\theta}{\partial x_3} \nonumber\\
	&\quad - \frac{1}{Re}\!\left(\frac{\partial^2 \hat{\mathfrak{G}}^{(v_3)}_\theta}{\partial x_1^2} + \frac{\partial^2 \hat{\mathfrak{G}}^{(v_3)}_\theta}{\partial x_2^2} + \frac{\partial^2 \hat{\mathfrak{G}}^{(v_3)}_\theta}{\partial x_3^2}\right),\label{eq:residual_3}\\[6pt]
	\hat{E}_{\theta 4} &= \frac{\partial \hat{\mathfrak{G}}^{(v_1)}_\theta}{\partial x_1} + \frac{\partial \hat{\mathfrak{G}}^{(v_2)}_\theta}{\partial x_2} + \frac{\partial \hat{\mathfrak{G}}^{(v_3)}_\theta}{\partial x_3}. \label{eq:residual_4}
\end{align}}
\end{subequations}
All spatial and temporal derivatives of the network outputs $\hat{\mathfrak{G}}_\theta^{(\cdot)}$ are computed via automatic differentiation (see Figure~\ref{fig:rotated}).

The composite loss is defined independently of the particular active branch subset. The dataset-construction section below makes explicit the bookkeeping used in implementation: fixed loss-target data determine which labels enter each supervised term, while the active branch-input subset determines which functions condition the operator. Supervised losses are evaluated on the training partition, whereas the physics residual can be sampled over the cardiac-cycle time interval because it requires no ground-truth labels.

\subsection{Dataset Construction}\label{sect:methodology:dataset_construction}

Having defined the proposed M$^{3}$PI-DeepONet and loss functions, we now describe how the CFD fields generated in Section~\ref{sect:methodology:GT_generation} are organized into a ground truth dataset used for training, validation, and testing purposes. This section covers data extraction into geometric strata, the spatio-temporal split, sparse supervision, and the distinction between loss-target data and branch-inputs.

\begin{figure}[htb]
	\centering	
	\includegraphics[scale=0.8]{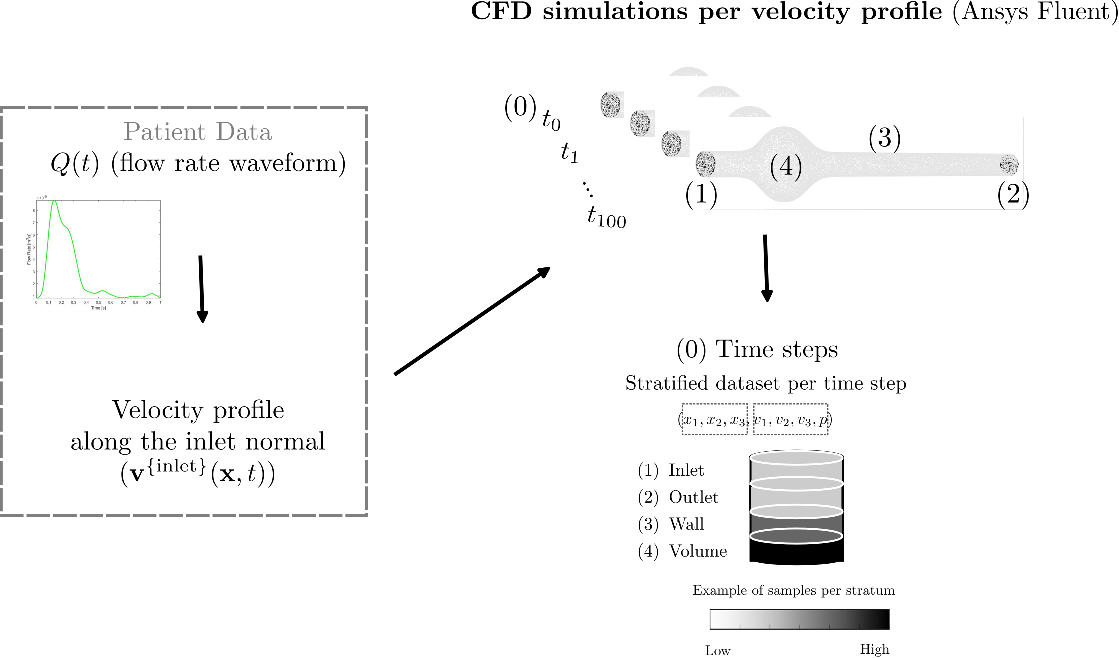}
    \caption{Overview of the dataset generation pipeline. Patient-specific flow rate waveforms $Q(t)$ are converted into time-dependent inlet velocity profiles $\boldsymbol{v}^{\{\text{inlet}\}}(\boldsymbol{x}, t)$. For each profile, a CFD simulation produces snapshots at discrete time steps $t_0, t_1, \ldots, t_{100}$. At every time step, the resulting data is organized into four geometrically distinct strata: (1) Inlet, (2) Outlet, (3) Wall, and (4) Volume.}
	\label{fig:data_generation}
\end{figure}

The raw CFD output underwent an automated extraction and preprocessing pipeline. At each recorded time step, the 3D mesh coordinates along with their associated velocity components and pressure values were extracted. To meet the specific requirements of the M$^{3}$PI-DeepONet training, in particular, the construction of distinct loss terms, the data was organized into four geometrically distinct strata corresponding to the \textit{Inlet}, \textit{Outlet}, \textit{Wall}, and internal \textit{Volume} regions of the domain. This stratification ensures that each velocity profile-dependent dataset provides direct access to the tuple $(x_1, x_2, x_3, v_1, v_2, v_3, p)$ for each stratum at every recorded time step (see Figure \ref{fig:data_generation}). Since the ground truth dataset is pre-computed and fixed, no dynamic resampling is performed during training.

\begin{figure}[htbp]
	\centering	
	\includegraphics[scale=0.65]{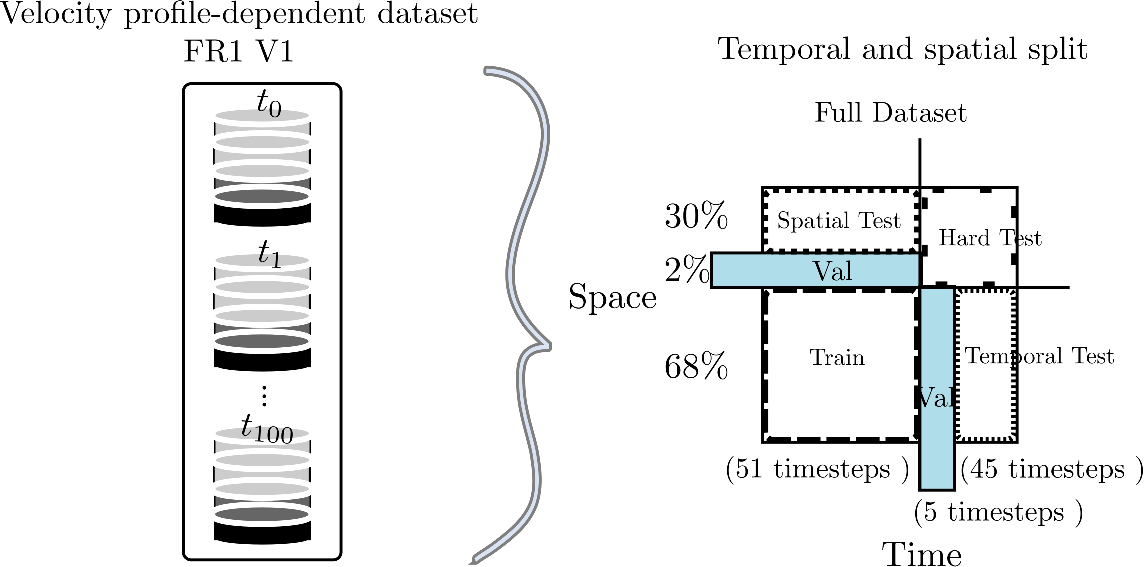}
	\caption{Spatio-temporal data splitting strategy for a single velocity profile-dependent dataset. The full set of 101 recorded time steps ($t_0$--$t_{100}$) is shown on the left. On the right, the dataset is partitioned along both axes: temporally into 51 training steps, a 5-step validation buffer, and 45 test steps; spatially into 68\% training, 2\% validation, and 30\% test. The model trains on the lower-left block. The remaining blocks define the three evaluation scenarios: \textit{Spatial Test}, \textit{Temporal Test}, and \textit{Hard Test}.}
	\label{fig:split_dataset}
\end{figure}

Because the M$^{3}$PI-DeepONet must learn a spatio-temporal operator, we employ a splitting strategy that partitions each velocity profile-dependent dataset along both the spatial and temporal dimensions simultaneously (see Figure \ref{fig:split_dataset}). Along the temporal axis, the initial condition at $t_0$, known exclusively at the boundaries (Inlet, Outlet, and Wall), is assigned to the training set; the remaining 100 recorded time steps ($t_1$ through $t_{100}$) are split into 50 additional training steps, 5 validation steps, and 45 testing steps, yielding 51 training time steps in total (including $t_0$). Along the spatial axis, the domain is partitioned into 68\% for training, 2\% for validation, and 30\% for testing, respectively.

M$^{3}$PI-DeepONet is trained exclusively on the intersection of these regions, designated as the \textit{Train} block: 68\% of the spatial points evaluated over the 51 training time steps. Similarly, model performance during training is monitored by two corresponding validation blocks: a \textit{Spatial Validation} block (2\% of unseen spatial locations evaluated over training time steps) and a \textit{Temporal Validation} block (training spatial locations evaluated over 5 unseen temporal steps). The remaining data blocks define three scenarios of increasing difficulty used for testing the model's generalization capabilities (see Figure \ref{fig:split_dataset}). The \textit{Spatial Test} evaluates predictions at unseen spatial locations for time steps seen during training, thereby assessing spatial generalization. The \textit{Temporal Test} evaluates predictions at training spatial locations for time steps not seen by the supervised training labels, assessing temporal generalization. Finally, the \textit{Hard Test} evaluates predictions at unseen spatial locations and time steps simultaneously.

It is important to note that the spatio-temporal splitting strategy constrains only the \textit{supervised} loss terms, namely $\mathcal{L}_{\text{data}}$, $\mathcal{L}_{\text{inlet}}$, $\mathcal{L}_{\text{outlet}}$, $\mathcal{L}_{\text{wall}}$, and $\mathcal{L}_{\text{ic}}$, whose evaluation points are drawn exclusively from the train block of the split. In contrast, the \textit{unsupervised} physics loss $\mathcal{L}_{\text{phy}}$ is not subject to this temporal partition, that is, at each training iteration, its collocation points are sampled with time values drawn uniformly at random between the minimum and maximum of the cardiac cycle. Because the physics residual requires no ground-truth labels, this unrestricted temporal sampling does not violate the spatio-temporal data splitting strategy and, crucially, allows the Navier-Stokes constraints to regularize the solution over the entire temporal domain, including the time steps reserved for the Temporal and Hard Test evaluations.


Additionally, within the designated \textit{Train} block, we intentionally use only a subset of the available data from each stratum, reflecting real-world scenarios in which complete measurements are rarely available across the entire domain. In addition, the stored fields can play two distinct roles: they may define supervision targets for the loss terms, or they may later be supplied as branch-input functions to the operator. We make this distinction explicit to avoid conflating what is used as training supervision with what is used as architectural input.

\textit{Loss-target data} (fixed). The supervision targets attached to each stratum are dictated by the physics of the problem and remain unchanged regardless of the branch inputs selected later in the architecture. Specifically, the boundary and initial-condition losses ($\mathcal{L}_{\text{inlet}}$, $\mathcal{L}_{\text{outlet}}$, $\mathcal{L}_{\text{wall}}$, $\mathcal{L}_{\text{ic}}$) supervise only the velocity field; the data loss $\mathcal{L}_{\text{data}}$ supervises both velocity and pressure at the sparse interior sensors introduced below; and the physics loss $\mathcal{L}_{\text{phy}}$ requires no targets at all, only $(\boldsymbol{x},t)$ collocation points. Pressure at the boundary regions is therefore not used as a supervision target. Adopting the tuple notation $(x_1, x_2, x_3, t, v_1, v_2, v_3, p)$, with $\text{NaN}$ flagging ``no target available'', the per-stratum target tuples are
\begin{align*}
\text{Inlet/Outlet:}\quad & (x_1, x_2, x_3, t,\, v_1, v_2, v_3,\, \text{NaN}), \\
\text{Wall (no-slip):}\quad & (x_1, x_2, x_3, t,\, 0, 0, 0,\, \text{NaN}), \\
\text{IC ($t = 0$):}\quad & (x_1, x_2, x_3, 0,\, v_1, v_2, v_3,\, \text{NaN}), \\
\text{Volume (collocation):}\quad & (x_1, x_2, x_3, t,\, \text{NaN}, \text{NaN}, \text{NaN},\, \text{NaN}), \\
\text{Sparse Data points:}\quad & (x_1, x_2, x_3, t,\, v_1, v_2, v_3,\, p).
\end{align*}

The NaN flags were consumed implicitly by the loss construction in Section~\ref{sect:adapting:deeponet:loss}, ensuring each loss term uses only the information physically meaningful at its stratum.

\begin{figure}[htbp]
	\centering	
	\includegraphics[scale=0.5]{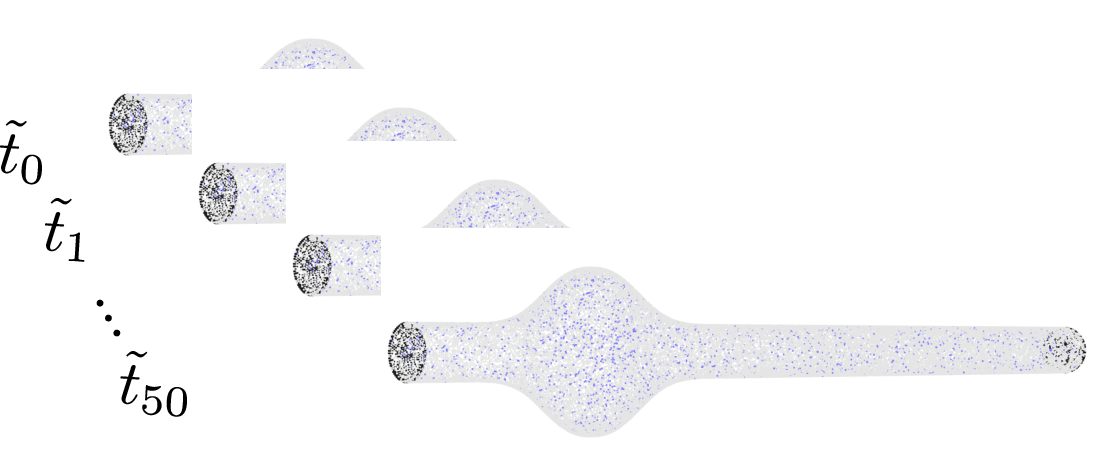}
	\caption{\textit{Data}: Mimicking an external sparse data source via random sampling within the Volume region. The temporal indices $\tilde{t}_i$ correspond directly to the 51 training time steps, and the $0.3\%$ spatial sampling is drawn once and reused identically across all these steps.}
	\label{fig:random_sample}
\end{figure}

The final row in the target tuples corresponds to the sparse supervised interior subset, denoted \textit{Data}. It consists of spatiotemporal coordinates within the \textit{Volume} region of the \textit{Train} block for which both velocity and pressure are retained as known targets. These points mimic an external sparse sensor source and enter training through the supervised data loss $\mathcal{L}_{\text{data}}$. The subset is constructed using a random spatial sampling of $0.3\%$ of the available volume points, where the identical set of spatial locations is reused across all training time steps, as illustrated in Figure~\ref{fig:random_sample}. We emphasize that this sparse interior subset is used exclusively as training supervision, through $\mathcal{L}_{\text{data}}$, and is never supplied to the network at inference. Predicting the fields for a new, unseen waveform therefore requires only the active branch inputs together with the query coordinates, and no interior measurement of the new case.

\textit{Branch-input data} (variable). The same stored fields may also serve as input functions for the branch networks. Unlike the fixed loss targets above, these inputs depend on the branch configuration selected in the architecture. The formal candidate set of branch inputs was introduced in Section~\ref{sect:methodology:network_arch:proposed_architecture}; here we only emphasize that changing the active branch inputs does not alter the supervision targets listed above. For example, inlet velocity can serve both as an inlet-loss target and as a branch-input source, whereas outlet pressure is used only as a possible branch-input source and wall velocity is used only as a loss target. 

This decoupling is important because scarce supervision remains fixed by the available measurements, whereas the choice of which boundary/initial functions to feed the operator is an architectural degree of freedom evaluated systematically in Section~\ref{sect:results:branch_selection}.

The detailed training protocol is provided in \ref{appendix:training_protocol}. In brief, all experiments use dimensionless inputs, an even/odd temporal instantiation of the spatio-temporal split, and a cross-waveform partition in which 12 of the 15 inlet profiles are used for training while three profiles are held out for full unseen-waveform testing. Unless otherwise stated, the optimizer settings, batching strategy, and loss-weight configurations follow \ref{appendix:training_protocol}.

\section{Results}\label{sect:results}

In this section, we conduct several numerical experiments to evaluate the performance of the proposed M$^{3}$PI-DeepONet in predicting velocity and pressure fields in the context of unsteady AAA simulations. To assess performance, we use the $L^2$-relative error as the principal accuracy metric. For a predicted scalar field $\hat{f}$ versus the CFD ground truth $f$, evaluated at $N$ spatio-temporal points $\{(\mathbf{x}_k, t_k)\}_{k=1}^{N}$, it is defined as
\begin{equation}
\mathcal{E}_{L^2}(f,\hat{f}) \;=\; \sqrt{\dfrac{\sum_{k=1}^{N}\bigl(f_k - \hat{f}_k\bigr)^2}{\sum_{k=1}^{N} f_k^{\,2}}}.
\label{eq:l2_rel_error}
\end{equation}

For qualitative inspection of spatial error distributions, we additionally use the pointwise \emph{absolute error} $|f_k - \hat{f}_k|$, which retains the physical units of the underlying field.

Hereafter, we will follow two conventions for handling errors: 
\begin{itemize}
    \item[ ] \emph{(i) Per-waveform.} For each individual waveform and each evaluation block (Spatial / Temporal / Hard / Full Test), $\mathcal{E}_{L^2}$ is computed on mini-batches of $10{,}000$ spatio-temporal coordinate points drawn uniformly at random from that block, and the per-waveform error is reported as the mean across batches.
    \item[ ] \emph{(ii) Across-waveforms.} The per-waveform means are then averaged over a chosen waveform set, typically the 3 unseen held-out waveforms (Full Test) or the 12 training waveforms, producing the headline ``$\mu \pm \sigma$'' reported in the result tables, whose standard deviation reflects the inter-waveform variability of the model's accuracy.
\end{itemize}

\subsection{Branch Input Selection}\label{sect:results:branch_selection}

To validate the multi-input formulation introduced in Section~\ref{sect:methodology:network_arch:proposed_architecture}, we first investigate which combination of physical inputs provides the operator with sufficient information to accurately predict the flow fields. For this purpose, we employ a standard Multi-Input Multi-Output PI-DeepONet (without the Modified architecture or the Aggregated Injection strategy) and systematically vary the number and type of branch inputs. All models share identical hyperparameters and use the unweighted loss setting ($\lambda_i = 1.0$ for all terms), following the protocol specified in \ref{appendix:training_protocol}.

\begin{table}[htbp]
	\centering
	\caption{Branch input study. $L^2$-relative errors (\%) for Magnitude of Velocity (MoV) and Pressure follow the across-waveforms conventions, averaging across the three held-out waveforms used for testing. Runtime denotes the total training wall-clock time. \#Params represents the total number of trainable parameters $|\theta|$ of the network.}
	\label{tab:results:branch_selection}
	\resizebox{\columnwidth}{!}{
		\begin{tabular}{clcccc}
			\toprule
			& & \multicolumn{2}{c}{\textbf{Full Test}} & & \\
			\cmidrule(lr){3-4}
			\textbf{No. Branches} & \textbf{Branch Inputs} & \textbf{MoV (\%)}  & \textbf{Pressure (\%)} & \textbf{Training Runtime (h)} & \textbf{\#Params}  \\
			\midrule
			1 & $v_2^{\{\text{inlet}\}}$  & 27.61 ± 0.27 & 12.16 ± 3.44 & 2.33 & 2.4990e+5 \\
			1 & $p^{\{\text{outlet}\}}$  & 26.39 ± 0.58 & 9.82 ± 2.76 & 2.38 & 2.4990e+5 \\
			1 & $v_2^{\{\text{ic}\}}$  & 14.20 ± 0.88 & 8.51 ± 0.84 & \textbf{2.3} & \textbf{2.2720e+5} \\
			\midrule
			2 & $v_2^{\{\text{inlet}\}}$,\; $p^{\{\text{outlet}\}}$  & 23.44 ± 0.89 & 11.94 ± 2.28 & 2.56 & 4.2860e+5 \\
			2 & $v_2^{\{\text{inlet}\}}$,\; $v_2^{\{\text{ic}\}}$  & \textbf{11.10 ± 0.23} & \textbf{9.17 ± 2.34} & 2.57 & 4.0590e+5 \\
			2 & $p^{\{\text{outlet}\}}$,\; $v_2^{\{\text{ic}\}}$  & 11.18 ± 0.27 & 9.26 ± 3.24 & 2.58 & 4.0590e+5 \\
			\midrule
			3 & $v_2^{\{\text{inlet}\}}$,\; $p^{\{\text{outlet}\}}$,\; $v_2^{\{\text{ic}\}}$  & 16.96 ± 0.77 & 9.52 ± 1.22 & 2.8 & 5.8460e+5 \\
			\bottomrule
		\end{tabular}
	}
\end{table}

\begin{figure}[htb]
	\centering
	\includegraphics[width=\columnwidth]{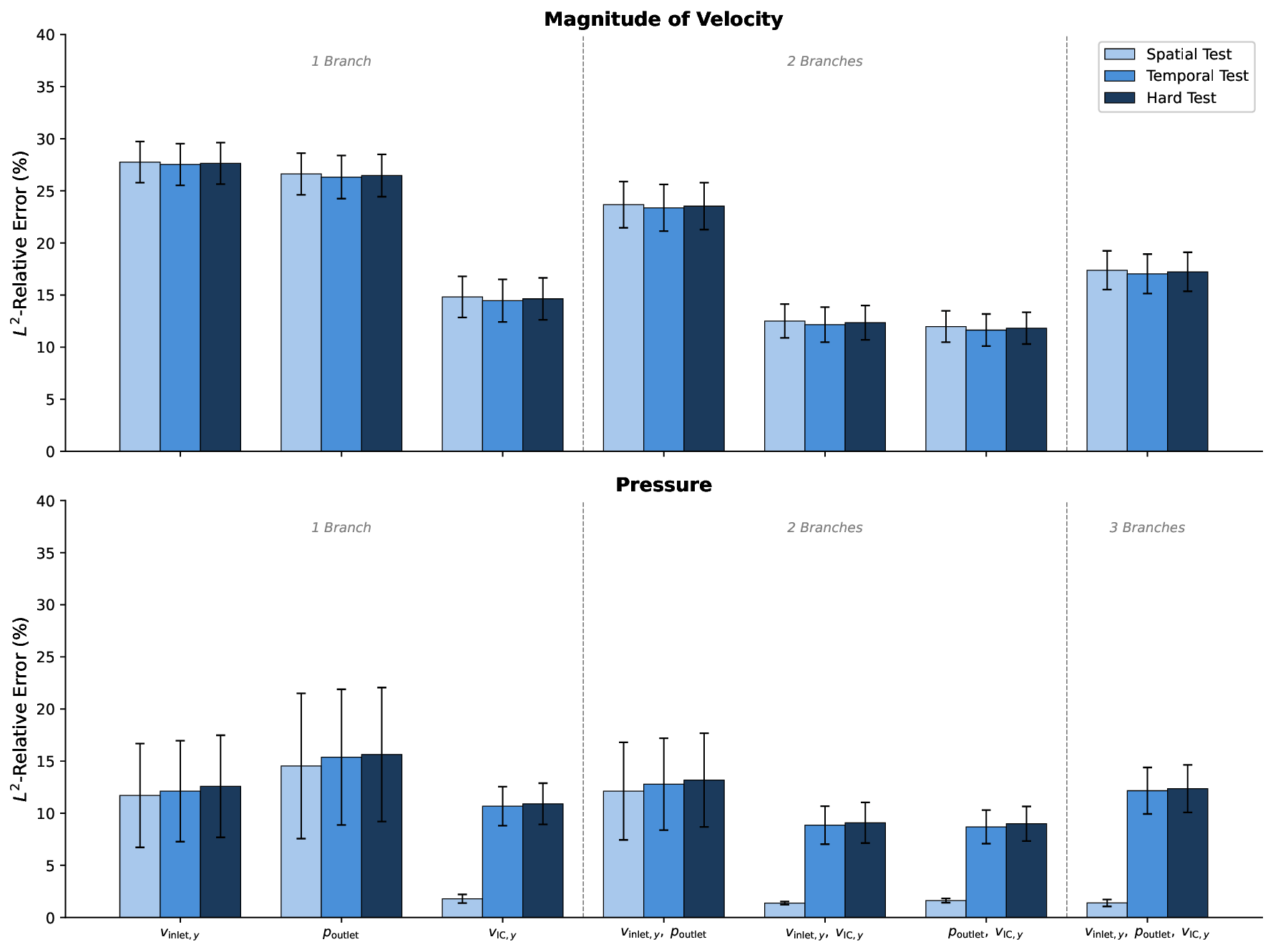}
	\caption{Impact of branch input selection on prediction accuracy. Top panel: magnitude of velocity; bottom panel: pressure. For each branch configuration, grouped bars show the $L^2$-relative error (\%) follow the across-waveform convention for the 12 training
    waveforms. In addition, the three test protocols (Spatial, Temporal, and Hard) are shown, with error bars indicating the standard deviation across waveforms. Configurations are separated by dashed vertical lines into single-branch (left), two-branch (center), and three-branch (right) groups. }
	\label{fig:results:branch_selection}
\end{figure}

Table~\ref{tab:results:branch_selection} reports the full test $L^2$-relative error (\%) for each configuration, and Figure~\ref{fig:results:branch_selection} provides a complementary visualization of the Spatial, Temporal, and Hard test errors across all branch combinations. It should be acknowledged that the $L^2$-relative error percentages reported here are non-negligible; what is most relevant in this analysis is therefore the behaviour of the different branch-input configurations rather than their accuracy. The number of trainable parameters (\#Params) is also explicitly reported to demonstrate that variations in predictive accuracy are driven by the physical relevance of the inputs, rather than merely reflecting differences in overall model capacity. Several observations emerge from these results.

First, among the single-branch models, the initial condition $v_2^{\{\text{ic}\}}$ yields markedly lower errors ($14.20\%$ MoV, $8.51\%$ pressure) than either the inlet velocity ($27.61\%$, $12.16\%$) or the outlet pressure ($26.39\%$, $9.82\%$). This indicates that the initial state carries the most informative signal for this operator. Second, all two-branch combinations that include $v_2^{\{\text{ic}\}}$ substantially outperform the single-branch models, confirming the complementarity of the physical inputs. Among the two-branch pairs, the combination ($v_2^{\{\text{inlet}\}}$, $v_2^{\{\text{ic}\}}$) achieves the best balance between velocity ($11.10\%$) and pressure ($9.17\%$) accuracy. Third, the three-branch model ($v_2^{\{\text{inlet}\}}$, $p^{\{\text{outlet}\}}$, $v_2^{\{\text{ic}\}}$) does not improve upon the best two-branch configuration and in fact exhibits a higher MoV error ($16.96\%$). This suggests that, without the Modified architecture together with the Aggregated Injection strategy, the standard MI-MO-PI-DeepONet struggles to effectively fuse three heterogeneous inputs, and the additional branch introduces optimization difficulties that outweigh the information gain. Notably, the pair ($v_2^{\{\text{inlet}\}}$, $p^{\{\text{outlet}\}}$), which omits the initial condition, performs worst among the two-branch models ($23.44\%$ MoV), further underscoring the critical role of $v_2^{\{\text{ic}\}}$. A further observation from Figure~\ref{fig:results:branch_selection} is the striking split in pressure errors across test protocols, configurations that include $v_2^{\{\text{ic}\}}$ achieve very low Spatial test pressure errors (below $2\%$), while their Temporal and Hard test pressure errors remain substantially higher ($\sim$9-11\%). This indicates that the initial condition branch provides strong spatial anchoring for pressure, but the temporal extrapolation of pressure remains a more challenging task.

These results establish two key findings that guide the remainder of this section: (i) the multi-input formulation is essential for capturing the coupled velocity-pressure dynamics, and (ii) the two-branch configuration ($v_2^{\{\text{inlet}\}}$, $v_2^{\{\text{ic}\}}$) already provides a competitive operator input while keeping the architecture compact.

\subsection{Ablation Study: Aggregated Injection and Training Techniques}\label{sect:results:ablation}

Having established the importance of multi-input formulation, we now investigate the contribution of the proposed modified architecture with Aggregated Injection strategy and complementary training techniques. To isolate each factor, we fix the branch configuration to two branches ($v_2^{\{\text{inlet}\}}$, $v_2^{\{\text{ic}\}}$) and systematically enable individual components, using the shared baseline settings and training-component definitions reported in \ref{appendix:training_protocol}. Table~\ref{tab:results:ablation} presents the results of this ablation study.

\begin{table}[htbp]
	\centering
	\caption{Ablation study isolating the contribution of each architectural and training component. All models use 2 branches ($v_2^{\{\text{inlet}\}}$, $v_2^{\{\text{ic}\}}$). Errors follow the across-waveforms conventions, averaging across the three held-out waveforms. 
	\textbf{AI} = Aggregated Injection (modified arch); \textbf{ED} = Exponential Decay; \textbf{DW} = data-weighted fixed loss setting ($\lambda_{\text{data}}=100$, all other weights equal to one); \textbf{KI} = Kaiming Initialization. }
	\label{tab:results:ablation}
	\resizebox{\columnwidth}{!}{
		\begin{tabular}{ccccccccc}
			\toprule
			&  & & & & \multicolumn{2}{c}{\textbf{Full Test}} & & \\
			\cmidrule(lr){6-7}
			\textbf{Config}  & \textbf{AI} & \textbf{ED} & \textbf{DW} & \textbf{KI} & \textbf{MoV (\%)} & \textbf{Pressure (\%)} & \textbf{Training Runtime (h)} & \textbf{\#Params} \\
			\midrule
			A1  & \ding{55} & \ding{55} & \ding{55} & \ding{55} & 11.10 ± 0.23 & 9.17 ± 2.34 & \textbf{2.58} & \textbf{4.0590e+5} \\
			A2  & \ding{51} & \ding{55} & \ding{55} & \ding{55} & 7.08 ± 1.19 & 10.65 ± 3.34 & 8.7 & 5.6940e+5 \\
			A3  & \ding{51} & \ding{51} & \ding{55} & \ding{55} & 6.62 ± 1.11 & 5.48 ± 1.05 & 8.6 & 5.6940e+5 \\
			A4  & \ding{51} & \ding{51} & \ding{51} & \ding{55} & 5.50 ± 0.75 & 5.35 ± 0.57 & 8.65 & 5.6940e+5 \\
			A5  & \ding{55} & \ding{51} & \ding{51} & \ding{51} & 10.92 ± 0.65 & 6.15 ± 1.48 & 3.26 &  4.0590e+5 \\
			A6  & \ding{51} & \ding{51} & \ding{51} & \ding{51} & \textbf{3.37 ± 0.06} & \textbf{4.25 ± 1.26} & 8.65 & 5.6940e+5 \\
			\bottomrule
		\end{tabular}
	}
\end{table}

The results reveal a clear hierarchy of contributions. The baseline MI-MO- PI-DeepONet without architectural enhancements (A1) serves as the reference point, with a MoV error of $11.10\%$ and a pressure error of $9.17\%$. A significant change is the introduction of the modified architecture with Aggregated Injection strategy (A1 $\rightarrow$ A2), which reduces the MoV error to $7.08\%$, a $36.21\%$ relative improvement, while pressure remains comparable ($10.65\%$). This result validates the central hypothesis of this work, that is, fusing branch representations prior to trunk injection allows the coordinate basis to adapt to the specific boundary conditions.

Adding exponential learning rate decay (A2 $\rightarrow$ A3) reduces the MoV and pressure errors to $6.62\%$ and $5.48\%$, nearly halving the pressure error relative to A2. This suggests that the modified architecture benefits from a refined optimization schedule that prevents late-stage overshooting. Fixing the loss weights (A3 $\rightarrow$ A4) further lowers the errors to $5.50\%$ and $5.35\%$, indicating that stable weighting reduces oscillations from adaptive balancing methods. Introducing Kaiming Uniform initialization (A4 $\rightarrow$ A6), further reduces errors to $\mathbf{3.37\%}$ MoV and $\mathbf{4.25\%}$ pressure, confirming the importance of variance-preserving initialization for the deep modified-layer architecture. 

However, when all techniques are activated except the modified architecture with Aggregated Injection (A5) the error in MoV increases to $10.92\%$ MoV, similar to baseline A1, showing that components work synergistically. Therefore, we adopt the full A6 configuration for the final model evaluation: physics-informed loss, modified architecture with Aggregated Injection, exponential decay, fixed weights, and Kaiming Uniform initialization.
\subsection{Final Model Performance}\label{sect:results:final_performance}

We now evaluate the fully optimized M$^{3}$PI-DeepONet, incorporating all beneficial techniques identified in the ablation study (see Config. A6). We compare three branch configurations to determine the optimal number of input functions (see Table \ref{tab:results:final_summary}). 
\begin{table}[htbp]
	\centering
	\caption{M$^{3}$PI-DeepONet performance for the optimized training configuration. Errors follow the across-waveforms conventions, averaging across the three held-out waveforms.}
	\label{tab:results:final_summary}
	\resizebox{\columnwidth}{!}{
		\begin{tabular}{clcccc}
			\toprule
			& & \multicolumn{2}{c}{\textbf{Full Test}} & & \\
			\cmidrule(lr){3-4}
			\textbf{Branches} & \textbf{Branch Inputs} & \textbf{MoV (\%)} & \textbf{Pressure (\%)} & \textbf{Training Runtime (h)} & \textbf{\#Params} \\
			\midrule
			2 & $v_2^{\{\text{inlet}\}}$,\; $v_2^{\{\text{ic}\}}$  &  \textbf{3.37 ± 0.06} & 4.25 ± 1.26 & \textbf{8.65} & \textbf{5.6940e+5} \\
			3 & $v_2^{\{\text{inlet}\}}$,\; $p^{\{\text{outlet}\}}$,\; $v_2^{\{\text{ic}\}}$  & 4.21 ± 1.28 & 4.69 ± 1.27 & 11.68 & 8.4600e+5 \\
			4 & $v_2^{\{\text{inlet}\}}$,\; $p^{\{\text{outlet}\}}$,\; $v_2^{\{\text{ic}\}}$,\; $p^{\{\text{ic}\}}$  & 3.63 ± 0.67 & \textbf{3.69 ± 1.08} & 14.48 & 1.0772e+6 \\
			\bottomrule
		\end{tabular}
	}
\end{table}

Once the full training recipe is applied, all three configurations become broadly competitive, but they expose different trade-offs. The two-branch pair $(v_2^{\{\text{inlet}\}},\, v_2^{\{\text{ic}\}})$ provides the most favorable accuracy-complexity balance, achieving the lowest MoV error ($3.37 \pm 0.06\%$), competitive pressure error ($4.25 \pm 1.26\%$), the shortest training runtime ($8.65$\,h), and the fewest trainable parameters among the optimized M$^{3}$PI-DeepONet models. The three-branch natural triple remains informative because its MoV error decreases from $16.96\%$ under the standard MI-MO-PI-DeepONet to $4.21\%$ with the full recipe, showing that Aggregated Injection and the other techniques working in synergy  substantially mitigates the multi-branch fusion difficulty identified in Table~\ref{tab:results:branch_selection}; however, it does not surpass the two-branch pair and requires additional parameters and training time. The four-branch extension including $p^{\{\text{ic}\}}$ achieves the lowest pressure error ($3.69 \pm 1.08\%$), but this gain comes with the largest model and longest runtime. We therefore interpret the four-branch result as a sensitivity check on an additional, partially dependent information channel rather than as evidence that more branches are uniformly beneficial. 

Notice that, on the one hand, the branch input $v_2^{\{\text{inlet}\}}$ can be obtained from the prescribed time-dependent velocity profile in the inlet region; however, on the other hand, the remaining branches inputs involved in the calculation in Table \ref{tab:results:final_summary} are partially ($v_2^{\{\text{ic}\}}$) or totally ($p^{\{\text{outlet}\}}$, $p^{\{\text{ic}\}}$) sampled from the retained CFD cycle, so these results on new unseen waveforms should be interpreted as a measurement-conditioned operator assessment rather than as a waveform-only surrogate evaluation. 

To provide a detailed view of the model's behavior across all waveforms, Table~\ref{tab:results:final_detailed} reports the per-waveform $L^2$-relative errors for the two-branch configuration $(v_2^{\{\text{inlet}\}},\, v_2^{\{\text{ic}\}})$.

\begin{table}[htbp]
	\centering
	\caption{Per-waveform $L^2$-relative error (\%) for the M$^{3}$PI-DeepONet with the two-branch configuration $(v_2^{\{\text{inlet}\}},\, v_2^{\{\text{ic}\}})$. Results are grouped by patient source and train/test assignment. The bottom three rows report the across-waveform ``$\mu \pm \sigma$'' over the training set, the test set, and all 15 waveforms.}
	\label{tab:results:final_detailed}
	\resizebox{\columnwidth}{!}{
		\begin{tabular}{ccccccccc}
			\toprule
			& & & \multicolumn{3}{c}{\textbf{Magnitude of Velocity}} & \multicolumn{3}{c}{\textbf{Pressure}} \\
			\cmidrule(lr){4-6} \cmidrule(lr){7-9}
			\textbf{Patient Group} & \textbf{Waveform} & \textbf{Split} & \textbf{Spatial Test} & \textbf{Temporal Test} & \textbf{Hard Test} & \textbf{Spatial Test} & \textbf{Temporal Test} & \textbf{Hard Test} \\
			\midrule
			FR1 & V1 & Training set & 4.07 & 4.01 & 4.04 & 0.33 & 4.09 & 4.43 \\
			FR1 & V3 & Training set & 2.74 & 2.69 & 2.72 & 0.25 & 3.88 & 4.26 \\
			FR1 & V4 & Training set & 3.22 & 3.16 & 3.20 & 0.27 & 3.99 & 4.36 \\
			FR1 & V5 & Training set & 2.76 & 2.71 & 2.74 & 0.24 & 3.89 & 4.26 \\
			\rowcolor{gray!15}
			FR1 & V2 & \textbf{Test set} & 3.24 & 3.19 & 3.22 & 0.73 & 4.04 & 4.38 \\
			\midrule
			FR2 & V1 & Training set & 2.90 & 2.88 & 2.90 & 0.19 & 6.58 & 7.33 \\
			FR2 & V2 & Training set & 3.34 & 3.30 & 3.33 & 0.21 & 6.64 & 7.38 \\
			FR2 & V4 & Training set & 2.91 & 2.89 & 2.91 & 0.20 & 6.55 & 7.29 \\
			FR2 & V5 & Training set & 3.79 & 3.73 & 3.77 & 0.24 & 6.68 & 7.42 \\
			\rowcolor{gray!15}
			FR2 & V3 & \textbf{Test set} & 3.10 & 3.06 & 3.09 & 0.22 & 6.61 & 7.35 \\
			\midrule
			FR3 & V1 & Training set & 3.45 & 3.38 & 3.42 & 0.36 & 3.47 & 3.71 \\
			FR3 & V3 & Training set & 2.78 & 2.73 & 2.76 & 0.31 & 3.38 & 3.62 \\
			FR3 & V4 & Training set & 2.59 & 2.56 & 2.58 & 0.32 & 3.36 & 3.60 \\
			FR3 & V5 & Training set & 3.06 & 3.01 & 3.05 & 0.35 & 3.44 & 3.69 \\
			\rowcolor{gray!15}
			FR3 & V2 & \textbf{Test set} & 4.62 & 4.52 & 4.57 & 1.04 & 3.46 & 3.72 \\
			\midrule
			\multicolumn{3}{c}{\textbf{Mean ± Std (12 Training sets)}} & 3.13 ± 0.44 & 3.09 ± 0.43 & 3.12 ± 0.43 & 0.27 ± 0.06 & 4.66 ± 1.40 & 5.11 ± 1.61 \\
			\multicolumn{3}{c}{\textbf{Mean ± Std (3 Test sets)}} & 3.65 ± 0.69 & 3.59 ± 0.66 & 3.63 ± 0.67 & 0.66 ± 0.34 & 4.70 ± 1.37 & 5.15 ± 1.58 \\
			\multicolumn{3}{c}{\textbf{Mean ± Std (All 15 sets)}} & 3.24 ± 0.54 & 3.19 ± 0.52 & 3.22 ± 0.53 & 0.35 ± 0.22 & 4.67 ± 1.39 & 5.12 ± 1.60 \\
			\bottomrule
		\end{tabular}
	}
\end{table}

The per-waveform analysis in Table~\ref{tab:results:final_detailed} reveals several noteworthy patterns. Velocity errors are remarkably consistent across all waveforms and patient groups, with Hard test MoV errors ranging from $2.58\%$ (FR3~V4) to $4.57\%$ (FR3~V2), and minimal degradation between Spatial, Temporal, and Hard test protocols. It is entirely expected that the lowest and highest errors across the dataset correspond to a training waveform and an unseen test waveform, respectively; the fact that both extremes belong to the FR3 group simply reflects the large intra-group amplitude variability of these specific profiles. This uniformity indicates that the operator has learned a robust velocity representation that generalizes well across both space and time.

Pressure predictions exhibit greater variability across patient groups. The FR1 and FR3 groups achieve Hard test pressure errors of approximately $3.6$--$4.4\%$, while the FR2 group exhibits consistently higher pressure errors of approximately $7.3$--$7.4\%$. This discrepancy can be attributed to the distinct temporal characteristics of the FR2 waveforms, which display sharper systolic peaks and more pronounced diastolic variations, creating steeper pressure gradients that are inherently more challenging for the network to resolve. A striking observation is the extremely low Spatial test pressure errors (below $1\%$ for all training waveforms), indicating that the model captures the spatial pressure distribution with high fidelity at training time steps, while pressure prediction at label-held-out temporal phases remains the most challenging aspect.

\begin{figure}[htbp]
	\centering
	\includegraphics[width=\columnwidth]{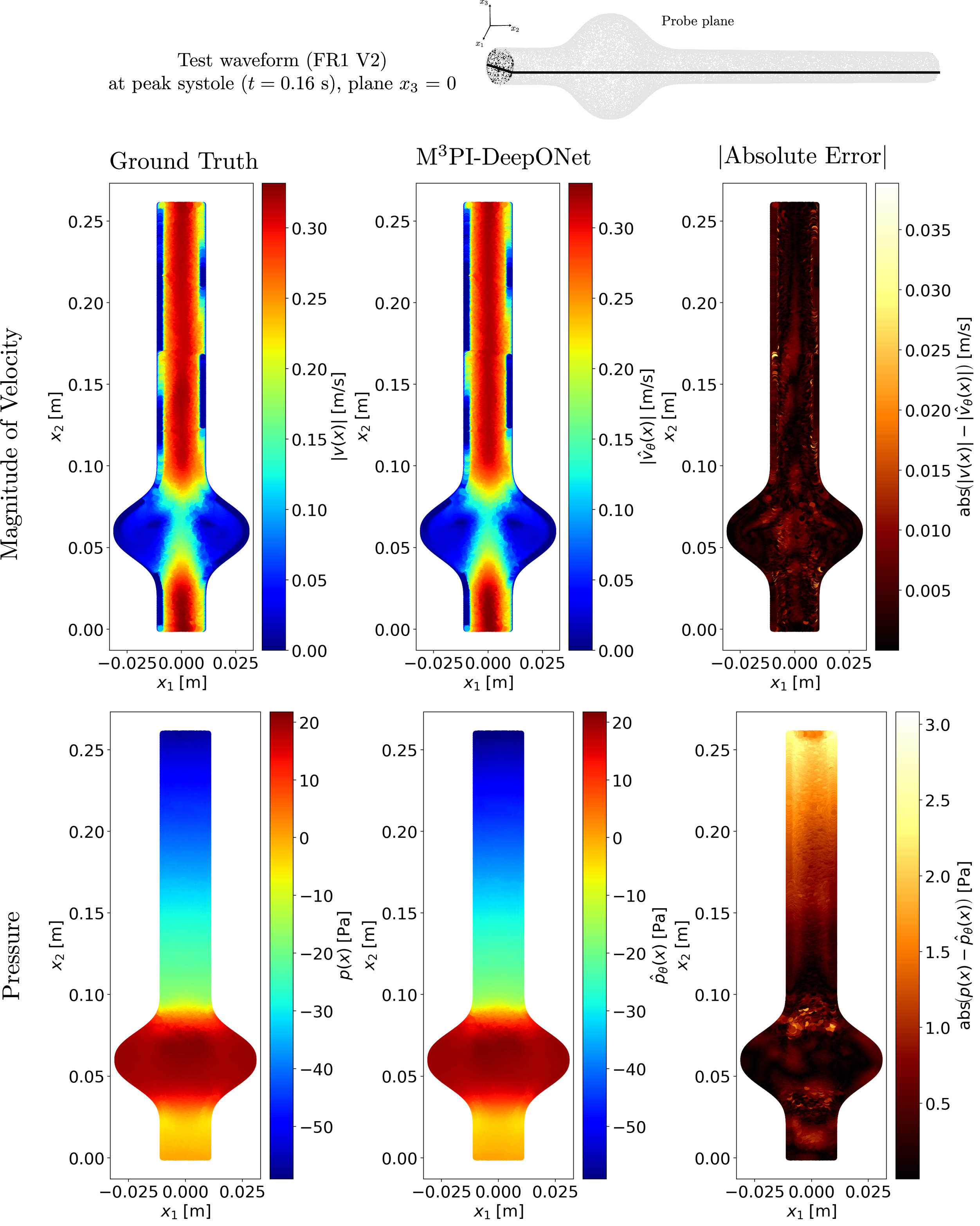}
	\caption{Qualitative comparison between the CFD ground truth (left), M$^{3}$PI-DeepONet prediction using the two-branch configuration $(v_2^{\{\text{inlet}\}},\, v_2^{\{\text{ic}\}})$ (center), and pointwise absolute error (right) for the held-out test waveform FR1$ $V2 at peak systole ($t = 0.16$\,s). Top row: magnitude of velocity; bottom row: pressure. }
	\label{fig:results:qualitative_fields}
\end{figure}

Notably, the three held-out test waveforms (shaded rows) achieve errors comparable to the training waveforms, confirming the operator's ability to generalize to entirely unseen waveform cases under their associated branch-input fields. The test-set mean MoV Hard error ($3.63\%$) is only slightly higher than the training-set mean ($3.12\%$), and the test pressure Hard error ($5.15\%$) is likewise close to the training value ($5.11\%$), demonstrating strong generalization. Among the test waveforms, the FR3~V2 test case exhibits the highest MoV error ($4.57\%$), while the FR2~V3 test case exhibits the highest pressure Hard error ($7.35\%$), consistent with the group-level trends observed in the training data.

Figure~\ref{fig:results:qualitative_fields} provides a qualitative comparison between the CFD ground truth and the two-branch M$^{3}$PI-DeepONet predictions for the held-out test waveform FR1 V2 at peak systole. FR1 V2 is selected as the representative case for the main text because it sits at \emph{intermediate} difficulty according to Table~\ref{tab:results:final_detailed}: its Hard MoV error ($3.22\%$) and Hard pressure error ($4.38\%$) lie between those of FR2 V3 (lowest MoV, highest pressure error) and FR3 V2 (highest MoV, lowest pressure error), so the displayed fields are not biased toward an unusually favourable test instance. The predicted velocity field accurately captures the key hemodynamic features, including the high-velocity jet entering the aneurysm bulge, the recirculation zone that forms along the dilated wall, and the near-zero boundary layer consistent with the no-slip condition. The pressure field faithfully reproduces the longitudinal pressure gradient along the vessel length, with the characteristic pressure drop across the aneurysm region. The absolute error maps confirm that the largest discrepancies are localized in high-gradient regions near the proximal and distal necks of the aneurysm, where the flow transitions sharply between the healthy vessel and the dilated segment. 

\begin{figure}[htbp]
	\centering
	\includegraphics[width=\columnwidth]{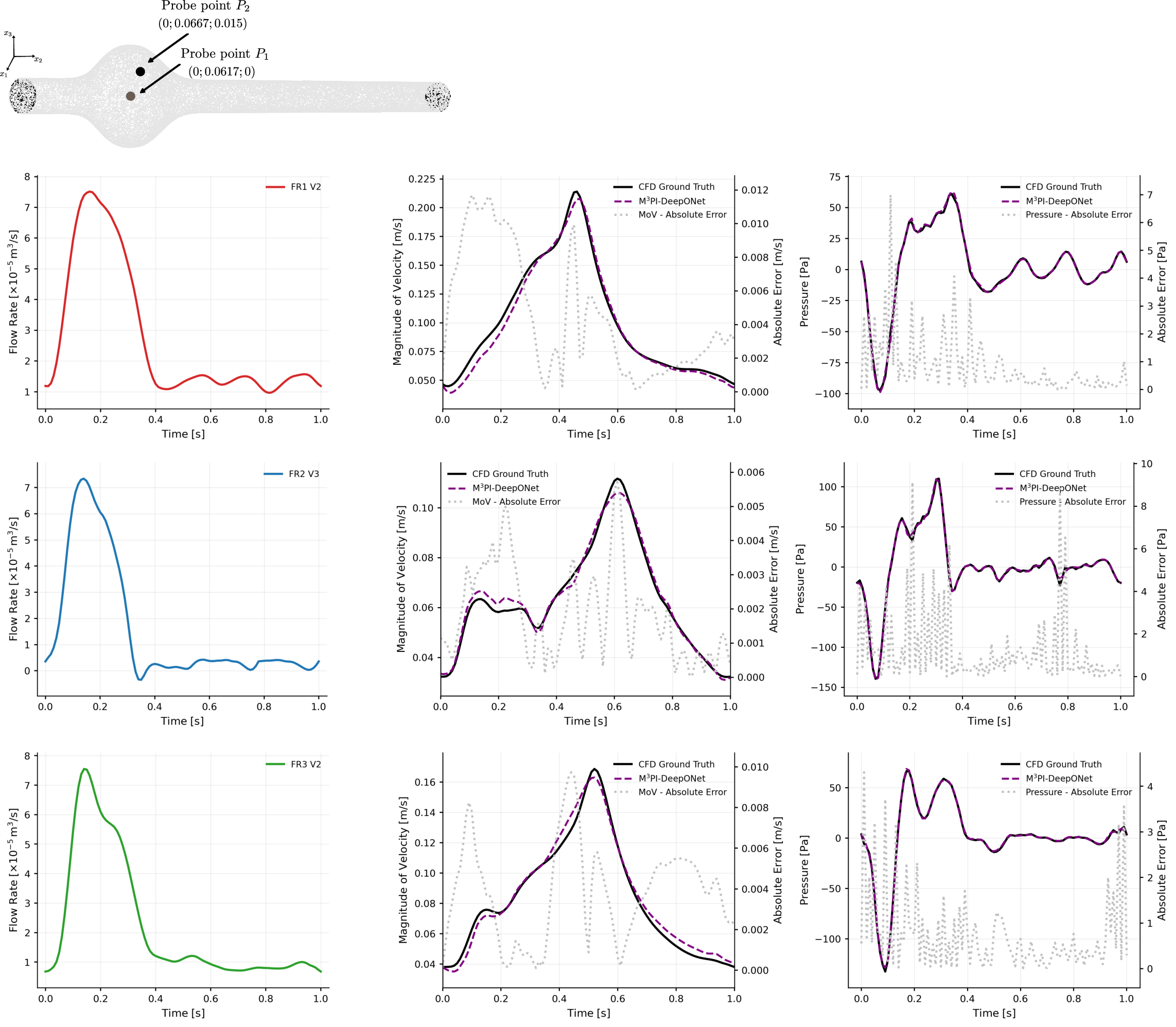}
	\caption{Temporal evolution of the predicted (dashed) and ground truth (solid) fields at probe point P$_1$, located at the center of the aneurysm bulge (coordinates $(0,\,0.0617,\,0)$), for the three held-out test waveforms, one per patient group (FR1$\cdot$V2, FR2$\cdot$V3, FR3$\cdot$V2). (a) Magnitude of velocity. (b) Pressure. The two-branch M$^{3}$PI-DeepONet accurately tracks both the systolic peak and diastolic recovery phases. Notice that the inset identifies both probe locations, P$_1$ and P$_2$, whereas the temporal curves shown here correspond only to P$_1$.}
	\label{fig:results:time_evolution}
\end{figure}

Figure~\ref{fig:results:time_evolution} shows the temporal evolution of velocity and pressure at probe point P$_1$, located at the geometric center of the aneurysm bulge, for the three held-out test waveforms using the two-branch configuration. This location lies inside the recirculation core, where the local field exhibits relatively damped temporal oscillations because the surrounding vortex structure acts as a low-pass filter on the cycle dynamics. The M$^{3}$PI-DeepONet predictions closely track the ground truth throughout the cardiac cycle at this point, including the rapid acceleration during systole and the gradual deceleration during diastole, with no discernible phase lag or systematic amplitude mismatch in velocity. Pressure predictions also follow the temporal dynamics faithfully, although slightly larger deviations are observed around the systolic peak, where the pressure gradient changes most rapidly.

The same evaluation was performed at probe point P$_2$, located off the symmetry axis at $(x_1,x_2,x_3)=(0.0,\,0.0667,\,0.015)$ and closer to the jet--recirculation interface. As indicated in the inset of Figure~\ref{fig:results:time_evolution}, P$_2$ therefore provides a complementary, more dynamically active sampling location. Because P$_1$ and P$_2$ yielded comparably small errors for all three held-out waveforms, only the P$_1$ curves are retained in Figure~\ref{fig:results:time_evolution} to avoid redundant plots, while the quantitative results for both probes are reported in Table~\ref{tab:results:probe_errors}.

\begin{table}[htbp]
	\centering
	\caption{Quantitative evaluation of the absolute errors at the two probe points shown in the inset of Figure~\ref{fig:results:time_evolution}: P$_1$ at the bulge centre, $(0,\,0.0617,\,0)$, and P$_2$ at the off-axis location $(0.0,\,0.0667,\,0.015)$. Results are reported for the three held-out test waveforms using the two-branch configuration.}
	\label{tab:results:probe_errors}
	\resizebox{\columnwidth}{!}{
	\begin{tabular}{lllcccc}
		\toprule
		& & & \multicolumn{2}{c}{\textbf{Velocity ($v$) [m/s]}} & \multicolumn{2}{c}{\textbf{Pressure ($p$) [Pa]}} \\
		\cmidrule(lr){4-5} \cmidrule(lr){6-7}
		\textbf{Probe} & \textbf{Patient} & \textbf{Test Waveform} & \textbf{MAE} & \textbf{Peak} & \textbf{MAE} & \textbf{Peak} \\
		\midrule
		P$_1$ (bulge centre) & FR1 & V2 & 0.0045 & 0.0117 & 0.852 & 6.994 \\
		P$_1$ (bulge centre) & FR2 & V3 & 0.0019 & 0.0057 & 1.253 & 9.136 \\
		P$_1$ (bulge centre) & FR3 & V2 & 0.0038 & 0.0097 & 0.730 & 4.304 \\
		\multicolumn{3}{l}{\textbf{Mean $\pm$ Std (P$_1$)}} & 0.0034 $\pm$ 0.0014 & 0.0091 $\pm$ 0.0031 & 0.945 $\pm$ 0.274 & 6.811 $\pm$ 2.421 \\
		\midrule
		P$_2$ (off-axis) & FR1 & V2 & 0.0036 & 0.0090 & 0.779 & 6.040 \\
		P$_2$ (off-axis) & FR2 & V3 & 0.0015 & 0.0054 & 1.234 & 9.499 \\
		P$_2$ (off-axis) & FR3 & V2 & 0.0027 & 0.0079 & 0.664 & 4.222 \\
		\multicolumn{3}{l}{\textbf{Mean $\pm$ Std (P$_2$)}} & 0.0026 $\pm$ 0.0011 & 0.0075 $\pm$ 0.0018 & 0.892 $\pm$ 0.301 & 6.587 $\pm$ 2.681 \\
		\bottomrule
	\end{tabular}}
\end{table}

Table~\ref{tab:results:probe_errors} summarizes the absolute velocity and pressure errors at both probes for the three held-out test waveforms, one per patient group. For each field $f \in \{v,\,p\}$, two complementary scalar metrics are computed over the $N_t$ probe samples $\{t_i\}_{i=1}^{N_t}$ spanning the cardiac cycle: the mean absolute error, $\mathrm{MAE}=\tfrac{1}{N_t}\sum_{i=1}^{N_t}\lvert f_i^{\text{pred}}-f_i^{\text{true}}\rvert$, and the peak absolute error, $\mathrm{Peak}=\max_i\lvert f_i^{\text{pred}}-f_i^{\text{true}}\rvert$, which captures the worst-case instantaneous deviation. The four probe-averaged summary values are slightly lower at P$_2$ than at P$_1$, confirming that the model performs comparably well at the two locations despite their different local flow dynamics.

For a high-level visual summary of the two-branch predictions across the whole cardiac cycle, phase-locked filmstrips of the velocity and pressure fields sampled at $14$ time instants spanning the early-systole, acceleration, peak-systole, deceleration, and diastolic phases are reported in \ref{appendix:filmstrip}.

\subsection{Computational Efficiency}\label{sect:results:computational_efficiency}

Finally, we assess the computational cost of the M$^{3}$PI-DeepONet relative to the reference CFD simulations. Because the CFD data generation involves both internal time integration and four-cycle periodization, Table~\ref{tab:results:computational_cost} reports the cost of one resolved retained cycle, which constitutes the saved dataset used to train and test the operator.

\begin{table}[htbp]
	\centering
	\caption{Computational cost comparison between the CFD solver and the M$^{3}$PI-DeepONet. \textsuperscript{*}\emph{Full DS inference} denotes evaluation over the retained dataset representation of the converged cardiac cycle, discretized into $101$ uniformly spaced snapshots.}
	\label{tab:results:computational_cost}
	\resizebox{\columnwidth}{!}{
		\begin{tabular}{llcl}
			\toprule
			\textbf{Method} & \textbf{Stage} & \textbf{Time} & \textbf{Hardware} \\
			\midrule
			CFD (ANSYS Fluent) & Resolved retained cycle & $\sim$ 12\,h & 13th Gen Intel(R) Core(TM) \\ &  & & i9-13900K (3.00 GHz) RAM 128 GB  \\
						\midrule
			M$^{3}$PI-DeepONet & Training (one-time) &  8\,h - 15\,h & NVIDIA RTX A6000 \\
			M$^{3}$PI-DeepONet & Full DS Inference\textsuperscript{*} (per waveform) & $\sim$ 20\,min & NVIDIA RTX A6000 \\
			\midrule
			\multicolumn{2}{l}{\textbf{Retained-cycle inference speedup}} & $\boldsymbol{\sim}$\textbf{36$\times$} & \\
			\bottomrule
		\end{tabular}
	}
\end{table}

The reported inference speedup of approximately $36\times$ is a conservative retained-cycle comparison, it compares the cost of resolving one converged CFD cardiac cycle ($\sim$12\,h) with the cost of evaluating the trained M$^{3}$PI-DeepONet on the retained 101-snapshot representation of that cycle ($\sim$20\,min). Both values refer to the same 101-snapshot output resolution, so the comparison is already normalized at equal output. Although the solver advances through 1000 internal time steps per cycle to ensure convergence, the $\sim$12\,h cost is governed by the time integration together with the writing of the 101 retained snapshots, and not by the number of internal steps.

While the one-time training cost of 8--15\,h (depending on the number of branches) is non-negligible, it is paid only once for a given training campaign. Beyond this, generating the CFD ground-truth database for the 12 training waveforms constitutes the dominant offline cost: each waveform requires the full four-cycle periodization run, of which only the retained cycle ($\sim$12\,h, dominated by snapshot writing) is stored at full temporal resolution, while the three warm-up cycles are considerably cheaper. This upfront cost is amortized over subsequent evaluations, for which a new waveform within the trained physiological range can be evaluated in $\sim$20\,min once the required branch inputs are available. 

\section{Discussion and conclusions}\label{sect:conclusions}

In this work, we introduced the M$^{3}$PI-DeepONet, a Modified Multi-Input Multi-Output Physics-Informed Deep Operator Network for predicting unsteady 3D hemodynamics in an idealized Abdominal Aortic Aneurysm geometry. The central methodological contribution is the \textbf{Aggregated Injection} strategy, which bridges two previously disjoint approaches in operator learning: the layer-wise gating mechanism of Wang et al. \cite{Wang2022Improved}, designed for single-input architectures, and the multi-branch MIONet topology of Jin et al. \cite{Jin2022MultipleInput}, which relies on static trunk basis functions. By computing a learned weighted fusion of the latent representations from all active input branches prior to trunk injection, the Aggregated Injection provides the trunk network with a unified physical context, transforming its coordinate basis from static to \textit{input-adaptive}. To the best of our knowledge, this is the first work to explicitly combine these two mechanisms in a multi-branch operator learning setting.

\textbf{Summary of key findings.}
\begin{enumerate}
    \item \textit{Branch input selection} (Section~\ref{sect:results:branch_selection}): Within the candidate set $\mathcal{S}$, the initial condition $v_2^{\{\text{ic}\}}$ proved the single most informative branch input. In a \emph{standard} MI-MO-PI-DeepONet, the pair $(v_2^{\{\text{inlet}\}},\, v_2^{\{\text{ic}\}})$ gave the best trade-off, while activating the three-branch natural triple actually \emph{degraded} velocity accuracy, a clear sign that, without an explicit fusion mechanism, the standard architecture cannot integrate heterogeneous inputs.

    \item \textit{Ablation study} (Section~\ref{sect:results:ablation}): The Aggregated Injection strategy provided the single largest improvement. Subsequent additions of exponential learning rate decay, the data-weighted fixed loss setting, and Kaiming Uniform initialization yielded cumulative gains, with the full configuration (A6) achieving $3.37\%$ MoV and $4.25\%$ pressure error. 

    \item \textit{Final model performance} (Section~\ref{sect:results:final_performance}): With the full training recipe (\ref{appendix:training_protocol}), the three-branch natural triple recovers from $16.96\%$ to $4.21\%$ MoV. That is a direct empirical confirmation that Aggregated Injection enables effective multi-branch fusion. Across the optimized configurations, the two-branch pair $(v_2^{\{\text{inlet}\}},\, v_2^{\{\text{ic}\}})$ provided the best accuracy-complexity trade-off, achieving the lowest MoV error with the fewest parameters and shortest training time. The four-branch extension including $p^{\{\text{ic}\}}$ achieves the lowest pressure error, but at increased computational cost.

     \item \textit{Computational efficiency} (Section~\ref{sect:results:computational_efficiency}): The trained operator achieves a conservative retained-cycle inference speedup of approximately $36\times$ compared to the reference CFD solver, reducing the prediction time for the retained 101-snapshot representation of a converged cardiac cycle from $\sim$12\,h to $\sim$20\,min. After the one-time offline costs of CFD dataset generation and training, each new waveform evaluation becomes orders of magnitude faster than a fresh CFD simulation once the required branch-conditioning inputs are available.
\end{enumerate}

\textbf{Interpretation and implications.}
The results suggest that the ``Input-Adaptive'' property conferred by the Aggregated Injection strategy plays an important role in resolving the coupled physics of aneurysm flows, where the solution, including recirculation zones, flow separation regions, vortex structures and pressure gradients, varies substantially with the driving waveform and outlet-pressure context. The per-waveform analysis further revealed that pressure prediction in the FR2 patient group, whose waveforms exhibit sharper systolic peaks and greater temporal variability, remains the most challenging aspect (Hard test errors of $\sim$7\%), suggesting that the model's temporal resolution is the primary bottleneck for the most complex physiological inputs.

From a clinical perspective, the conservative $36\times$ retained-cycle inference speedup moves the framework closer to the goal of rapid hemodynamic assessment. In such a setting, the operator would be most useful if the conditioning inputs could be supplied from clinically accessible measurements or inexpensive auxiliary models, allowing a clinician to explore the effect of varying blood pressure waveforms on wall shear stress and pressure distributions without the prohibitive cost of repeated CFD simulations. The operator learning paradigm is well-suited to this scenario, but the present study demonstrates the conditional reconstruction step rather than the complete prospective clinical pipeline.

These clarifications also sharpen the contribution of the present work. The generation of the CFD ground-truth database remains the dominant offline bottleneck: high-fidelity Fluent simulations are needed to train, validate, and test the operator, and in the present benchmark they also provide some of the branch-conditioning quantities. The contribution of M$^{3}$PI-DeepONet is therefore not to eliminate all CFD dependence at this stage, but to demonstrate that, once a trained operator and the required branch inputs are available, the full unsteady three-dimensional velocity and pressure fields can be reconstructed rapidly from partial physical information.

For a practical use case, the two-branch configuration $(v_2^{\{\text{inlet}\}},\,v_2^{\{\text{ic}\}})$ retained for the final model provides the clearest interpretation. For a new waveform on the same idealized AAA geometry, the first step is to confirm that the waveform lies within the physiological range covered by the training set, in terms of both peak inlet velocity, or equivalently Reynolds number, and temporal shape. The minimum physical information that would then need to be provided to the inference pipeline is: the inlet streamwise velocity branch $v_2^{\{\text{inlet}\}}$, constructed from the new flow-rate waveform $Q(t)$ and sampled at the fixed inlet branch sensors; the periodic initial-state branch $v_2^{\{\text{ic}\}}$, sampled at the fixed initial-condition branch sensors at the selected $t=0$ phase (only its outlet trace at $t=0$ adds information beyond $v_2^{\{\text{inlet}\}}$; see Section~\ref{sect:methodology:network_arch:proposed_architecture}); and the same geometry, coordinate system, and temporal phase convention used during training. The dimensionless representation does not require an additional measurement: for the same geometry and fluid properties, the inlet radius $R$, density $\rho_f$, viscosity $\mu_f$, and cardiac period $\mathscr{T}$ are fixed, while the velocity scale $V$ is computed from the maximum inlet velocity of the new waveform, as in \ref{appendix:dimensionless}. Once the branch arrays are provided in the trained branch-sensor ordering and scaled consistently, the M$^{3}$PI-DeepONet can be queried at the desired spatial points and time instants to recover $(v_1,v_2,v_3,p)$.

\textbf{Limitations.}
Several simplifying assumptions in the present framework should be acknowledged. First, the operator was trained and tested on a single idealized abdominal aortic aneurysm (AAA) geometry; consequently, generalization to patient-specific anatomies with varying sac shape, curvature, asymmetry, and bifurcation topology remains to be demonstrated. Second, the vessel wall was modeled as rigid, so fluid--structure coupling (FSI) and the associated wall deformation were not represented. This approximation may not be too restrictive when considering AAA because this pathology stiffens the arterial wall. Third, the most accurate configurations in the present work rely on branch inputs, particularly $v_2^{\{\text{inlet}\}}$ and $v_2^{\{\text{ic}\}}$, that are partially known a priori in a new clinical case. The first is always available but the second is available here because it has been extracted from the retained fourth CFD cycle (more precisely, only its outlet trace at $t=0$ stems from CFD, since its inlet part follows the prescribed waveform and its wall part vanishes by no-slip), and therefore the reported results should be interpreted as a measurement-conditioned benchmark. Fourth, the training dataset comprises 15 inlet waveforms, a modest diversity that, while sufficient for proof-of-concept, may not span the full range of physiological variability encountered in clinical populations. Finally, the hyper-parameters were manually selected from commonly used values in the literature rather than systematically optimized, and all results are based on a single random seed; a more rigorous evaluation with multiple seeds and automated hyperparameter search would strengthen the statistical robustness of the reported errors.

\textbf{Future work.}
Several directions are envisioned to address the limitations identified above and to extend the applicability of the framework. First, the operator could be extended to accept geometry as an additional input, for instance by encoding the vessel shape through a dedicated branch using signed distance functions or graph-based representations. This would enable predictions across different patient-specific anatomies without requiring retraining. A second important direction concerns the definition of prospective branch inputs for a fully predictive CFD-free deployment. Depending on which branches are activated, this would involve constructing $v_2^{{\text{inlet}}}$ from the prescribed or measured $Q(t)$, estimating $p^{{\text{outlet}}}$ from pressure measurements or a 0D/1D circulation model, for instance a Windkessel model driven by the outlet flow rate measured from patient imaging, and approximating $v_2^{{\text{ic}}}$ through a reduced-order spin-up, a previous-cycle measurement, the velocity components measured at the boundaries at the initial instant, or a learned initialization model. Further improvements could also be achieved by incorporating compliant wall mechanics, either by coupling the operator with a structural model through fluid-structure interaction or by including wall displacement as an additional output field, thereby increasing the physiological fidelity of the predictions. In addition, replacing the Newtonian constitutive model with a generalized Newtonian formulation, such as Carreau-Yasuda, within the physics-informed loss would allow the model to capture shear-thinning effects without requiring additional labeled data. From an architectural perspective, incorporating a random Fourier feature embedding \cite{Tancik2020,Wang2021FourierF} at the trunk input could mitigate the spectral bias of neural networks toward low-frequency functions and potentially improve the resolution of sharp gradients characteristic of pulsatile hemodynamics, particularly in the high-gradient regions near the aneurysm necks where the largest prediction errors were observed. Self-adaptive loss balancing techniques could also be explored to dynamically adjust the relative weighting of the loss terms during training. Finally, robustness and scalability should be assessed more rigorously by conducting multiple training runs with varying random seeds and by applying automated hyper-parameter optimization through meta-learning or Bayesian search \cite{Chelsea2017,Wang2023Long}. Transfer learning strategies could further reduce the training cost associated with adapting the operator to new geometries or flow regimes.

\section*{Data and code availability}
The data and code that support the findings of this study are available from the corresponding author upon reasonable request.

\section*{CRediT authorship contribution statement}

\textbf{OLCG:} Conceptualization, Methodology, Software, Visualization, Formal analysis, Writing -- original draft. \textbf{BG:} Funding
acquisition, Conceptualization, Methodology, Formal analysis, Writing -- review \& editing. \textbf{VD:} Funding acquisition, Conceptualization, Methodology, Formal analysis, CFD data base constitution, Writing -- review \& editing.

\section*{Declaration of competing interest}

The authors declare that they have no known competing financial interests or personal relationships that could have
appeared to influence the work reported in this paper.

 \section*{Acknowledgements}\label{Acknowledgements}
This work received support from French government under the France 2030 investment plan, as a part of the Initiative d'Excellence d'Aix-Marseille Université - A*MIDEX, AMX-21-RID-022. The authors thank ANSYS for support.

\appendix

\newpage
\section{Training Protocol}\label{appendix:training_protocol}

\textit{Dimensionless approach.} The CFD ground truth is generated and stored in dimensional variables. Before training, coordinate, velocity, and pressure quantities supplied to the network are scaled using the same characteristic quantities used in the dimensionless analysis (see \ref{appendix:dimensionless}). Dimensional fields are recovered at inference by applying the inverse transformations.

\textit{Data export and storage.} The raw simulation data for each velocity profile initially occupied approximately 17.9\,GB. To streamline the machine-learning pipeline, an automated export routine converted the retained ANSYS Fluent results into serialized Python pickle files (\texttt{.pkl}), reducing the storage footprint by approximately $3.4\times$ to 5.24\,GB per profile. This extraction required roughly 1\,hour of computation time per velocity profile.

\textit{Experimental temporal split.} For all experiments reported in this work, the general temporal partition described in Section~\ref{sect:methodology:dataset_construction} was instantiated using an even/odd split. The \textit{Train} block used the even-indexed snapshots $t_0,t_2,\ldots,t_{100}$, where $t_{100}$ was retained only as the periodic closure of $t_0$. The Temporal Validation, Temporal Test, and Hard Test blocks were drawn from the odd-indexed snapshots $t_1,t_3,\ldots,t_{99}$.

\textit{Network sizing and optimizer.} Unless otherwise stated, all sub-networks (branches and trunk) are standard MLPs with 4 hidden layers of 100 neurons each, using the $\tanh$ activation function. All models were trained for $200{,}000$ iterations using the Adam optimizer \cite{Kingma2014Adam} with a learning rate of $\eta = 10^{-3}$ ($\beta_1 = 0.9$, $\beta_2 = 0.999$). In the baseline configuration, network weights and biases are initialized using the default Xavier (Glorot) Uniform scheme. As part of the ablation study (Section~\ref{sect:results:ablation}), two additional training enhancements are introduced: (i) an exponential learning rate decay with a decay rate of $0.95$ every $3{,}000$ iterations, and (ii) a Kaiming Uniform initialization \cite{He2015Kaiming,Chen2025BRDR}, where for each linear layer with $n_{\text{in}}$ input features, all parameters are drawn from $\mathcal{U}(-b,\,b)$ with $b = 1/\sqrt{n_{\text{in}}}$.

\textit{Implementation and hyperparameter selection.} The models were implemented in PyTorch and trained on a locally hosted workstation equipped with an NVIDIA RTX A6000 GPU with 48\,GB of memory. Hyperparameters were manually adjusted based on values commonly used in the literature, and systematic optimization is left to future work (see Limitations, Section~\ref{sect:conclusions}). All experiments used a fixed random seed of $42$ for reproducibility. The validation blocks defined in Section~\ref{sect:methodology:dataset_construction} were used to monitor training behaviour and are not included in the reported test metrics.

\begin{figure}[htbp]
	\centering
	\includegraphics[scale=0.4]{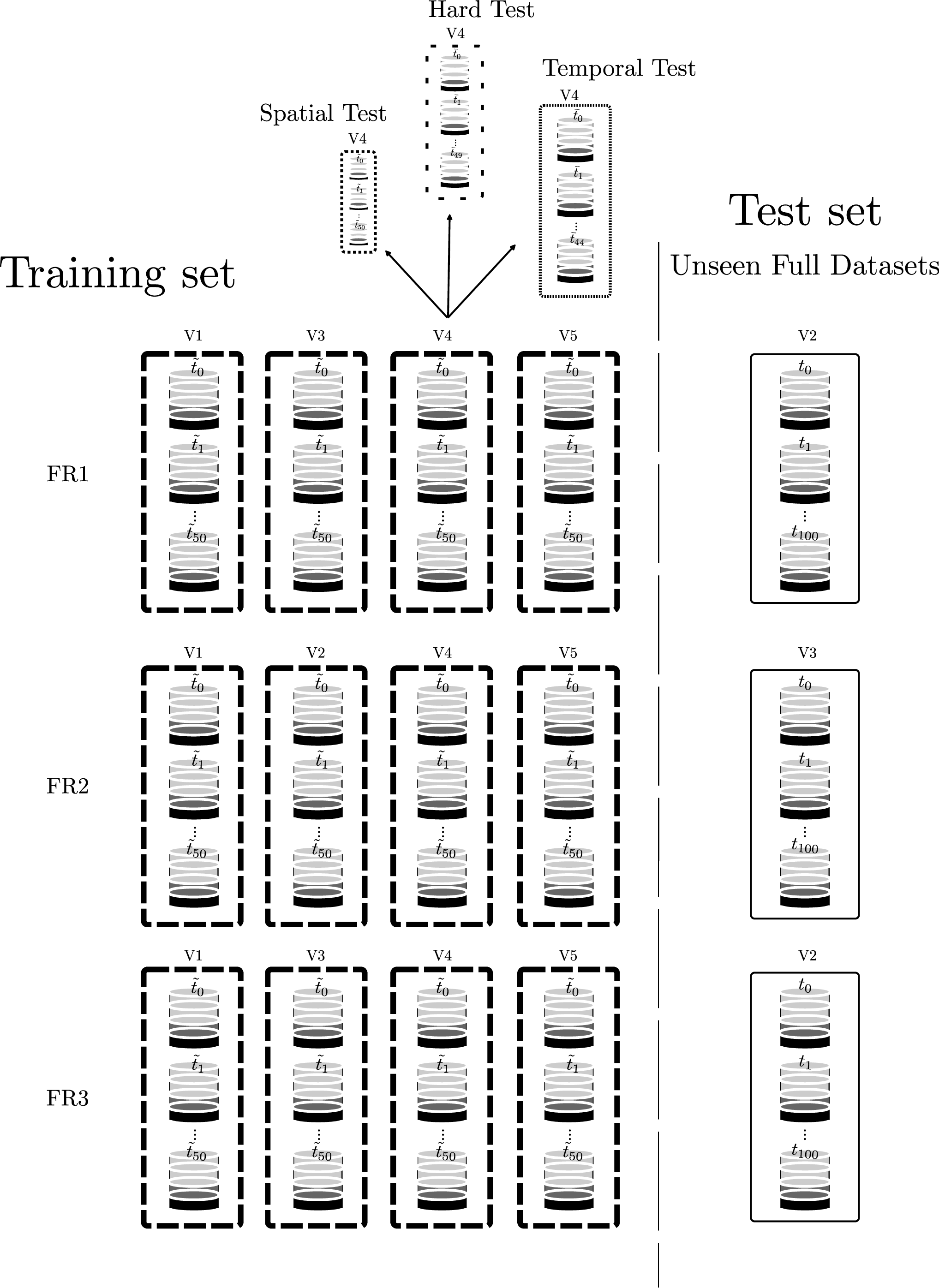}
	\caption{Cross-waveform dataset splitting process. The 15 velocity profiles are organized into three patient groups (FR1, FR2, FR3), each containing five variants (V1-V5). One specific variant from each group (V2 for FR1 and FR3, and V3 for FR2) is held out as an entirely unseen test set (3 profiles), evaluating the operator's generalization to new waveforms. The remaining 12 profiles form the training set, within which the spatio-temporal split of Figure~\ref{fig:split_dataset} applies.}
	\label{fig:train_test_split}
\end{figure}

\textit{Cross-waveform partitioning.} Beyond the within-waveform spatio-temporal split of Section~\ref{sect:methodology:dataset_construction}, the operator-learning premise requires evaluating generalisation \emph{across} waveforms. To this end, the 15 velocity profiles (3 patient groups $\times$ 5 variants each) are split into a training set of 12 profiles and a held-out test set of 3 profiles (one specific variant from each group: V2 for FR1 and FR3, and V3 for FR2), as illustrated in Figure~\ref{fig:train_test_split}. Within each of the 12 training profiles, the spatio-temporal splitting strategy of Section~\ref{sect:methodology:dataset_construction} applies, providing internal Spatial Test, Temporal Test, and Hard Test evaluation blocks. The three held-out test profiles constitute entirely unseen full datasets, testing the operator's ability to generalize to new waveforms within the physiological range covered by the training data.

\textit{Profile-level batching.} Rather than processing all 12 training profiles simultaneously, a random subset of 6 profiles is selected at each iteration to form the training batch. This stochastic sampling across the parameter space promotes generalization and reduces memory requirements while ensuring that the operator is exposed to diverse waveform conditions throughout training.

\textit{Point-level mini-batching.} To evaluate the components of the composite loss function (defined in Section~\ref{sect:adapting:deeponet:loss}) efficiently within each selected profile, we employ a decoupled mini-batching strategy. At each training iteration, subsets of spatio-temporal points are sampled uniformly at random from each boundary region within the \textit{Train} block to enforce boundary and initial conditions. The sampling fractions below were established empirically as hyperparameters to achieve an optimal trade-off between GPU memory constraints and predictive performance: we sample $2\%$ of the available Inlet spatio-temporal points, $2\%$ of the Outlet points, $0.02\%$ of the Wall points, and $2\%$ of the Initial Condition points. For the supervised data loss, $1\%$ of the sparse \textit{Data} points are sampled. Separately, a distinct mini-batch of exactly 1024 spatio-temporal collocation points is drawn to evaluate the physics residual loss. By processing only these small, randomized fractions of the massive computational dataset at each iteration, this approach substantially reduces the overall memory footprint, while still ensuring that the network robustly explores the entire spatio-temporal domain across training epochs.

\textit{Loss-term weighting.} No adaptive loss balancing is used in this work. Instead, we compare two fixed weighting settings for the composite loss (Eq.~\eqref{eq:loss_pideeponet}, defined in Section~\ref{sect:adapting:deeponet:loss}). In the unweighted setting, all terms are assigned unit weight: $\lambda_{\text{data}} = \lambda_{\text{inlet}} = \lambda_{\text{outlet}} = \lambda_{\text{wall}} = \lambda_{\text{ic}} = \lambda_{\text{phy}} = 1$. In the data-weighted setting, the sparse supervised data term is emphasized by setting $\lambda_{\text{data}} = 100$, while all remaining weights stay equal to one. The branch-selection study and the first ablation configurations use the unweighted setting, while the data-weighted setting is introduced explicitly in the ablation study and retained in the final model evaluation.

\newpage
\section{Dimensionless form of NSE for the unsteady AAA idealized model} \label{appendix:dimensionless}
In this section, we derive the dimensionless form of the unsteady Navier-Stokes equations \eqref{momentum3D}--\eqref{mass3D}. For this purpose, we choose constant parameters that are representative of the physical problem: the maximum inlet velocity ($V$), the inlet radius ($R = R_{\text{inlet}}$), and the angular frequency of the pulsatile flow ($\omega = 2\pi / \mathscr{T}$, where $\mathscr{T}$ is the cardiac period). We normalize the physical quantities as follows,
\begin{align}
    & \boldsymbol{x}^{*} = \frac{\boldsymbol{x}}{R}, \quad t^{*} = \omega\, t, \quad \boldsymbol{v}^{*} = \frac{\boldsymbol{v}}{V}, \quad p^{*} = \frac{p}{\rho_f V^2}. \label{eq:appendix:normalization}
\end{align}

Notice that, from Eq.~\eqref{eq:appendix:normalization} we can derive that,
\begin{align}
    & \nabla^{*} = R\, \nabla, \quad (\nabla^{*})^2 = R^2\, \nabla^2, \quad \frac{\partial}{\partial t} = \omega \frac{\partial}{\partial t^*}. \label{eq:appendix:nabla}
\end{align}

Based on Eqs.~\eqref{eq:appendix:normalization}-\eqref{eq:appendix:nabla}, we rewrite Eqs.~\eqref{momentum3D}-\eqref{mass3D} in terms of the dimensionless variables. Introducing the kinematic viscosity $\nu_f = \mu_f / \rho_f$, the substitution yields,
\begin{subequations}\label{eq:appendix:substituted}
\begin{align}
    & \omega V \frac{\partial \boldsymbol{v}^*}{\partial t^*}  + \frac{V^2}{R}(\boldsymbol{v}^* \cdot \nabla^*)\boldsymbol{v}^* =  -\frac{V^2}{R}\nabla^* p^* + \frac{V\,\nu_f}{R^2} (\nabla^*)^2 \boldsymbol{v}^*, \label{eq:appendix:substituted:a}\\
    &  \nabla^* \cdot \boldsymbol{v}^* = 0. \label{eq:appendix:substituted:b}
\end{align}
\end{subequations}

At this point, if we multiply Eq.~\eqref{eq:appendix:substituted:a} by $\displaystyle{\frac{R}{V^2}}$ on both sides, we obtain
\begin{subequations}\label{eq:appendix:simplified}
\begin{align}
    & \frac{\omega R}{V} \frac{\partial \boldsymbol{v}^*}{\partial t^*}  + (\boldsymbol{v}^* \cdot \nabla^*)\boldsymbol{v}^* =  -\nabla^* p^* + \frac{\nu_f}{R\, V} (\nabla^*)^2 \boldsymbol{v}^*, \\
    &  \nabla^* \cdot \boldsymbol{v}^* = 0.
\end{align}
\end{subequations}

Thus, the Reynolds number $\displaystyle{Re = \frac{R\, V}{\nu_f}}$ and the Womersley number $\displaystyle{\alpha^2 = \frac{\omega\, R^2}{\nu_f}}$ appear naturally in Eq.~\eqref{eq:appendix:simplified}, leading to the dimensionless form of the unsteady Navier-Stokes equations,
\begin{subequations}\label{eq:appendix:dimensionless_nse}
\begin{align}
    & \frac{\alpha^2}{Re} \frac{\partial \boldsymbol{v}^*}{\partial t^*}  + (\boldsymbol{v}^* \cdot \nabla^*)\boldsymbol{v}^* =  -\nabla^* p^* + \frac{1}{Re} (\nabla^*)^2 \boldsymbol{v}^*, \label{eq:appendix:dimensionless_momentum}\\
    &  \nabla^* \cdot \boldsymbol{v}^* = 0. \label{eq:appendix:dimensionless_mass}
\end{align}
\end{subequations}

Unsteady case introduces the temporal acceleration term $\frac{\alpha^2}{Re} \frac{\partial \boldsymbol{v}^*}{\partial t^*}$, whose magnitude relative to the convective term is governed by the Womersley number $\alpha$. When $\alpha \gg 1$, the oscillatory inertia dominates over viscous effects within each cycle, which is the regime relevant to large-artery pulsatile flows.

Table~\ref{tab:appendix:parameters} summarizes the physical parameters and the resulting dimensionless numbers for the 15 flow rate waveforms used in this study. Since all waveforms share the same geometry ($R = 0.010065\;\text{m}$), fluid properties ($\rho_f = 1060\;\text{kg/m}^3$, $\mu_f = 0.00399\;\text{kg/(m\,s)}$), and cardiac period ($\mathscr{T} = 1\;\text{s}$), the Womersley number $\alpha \approx 13.0$ is identical across all cases. The Reynolds number varies with the peak inlet velocity of each waveform.

\begin{table}[htbp]
	\centering
	\caption{Physical parameters and dimensionless numbers for the 15 waveforms.}
	\label{tab:appendix:parameters}
	\resizebox{10cm}{!}{%
		\begin{tabular}{ccccc}
			\toprule
			\textbf{Patient Group} & \textbf{Waveform} & $V$ [m/s] & $Re$ & $\alpha$ \\
			\midrule
			FR1 &  V1 & 0.388 & 1038 & 13.0 \\
			FR1 &  V2 & 0.331 & 884 & 13.0 \\
			FR1 &  V3 & 0.365 & 976 & 13.0 \\
			FR1 &  V4 & 0.340 & 910 & 13.0 \\
			FR1 &  V5 & 0.281 & 751 & 13.0 \\
			\midrule
			FR2 &  V1 & 0.346 & 926 & 13.0 \\
			FR2 &  V2 & 0.301 & 805 & 13.0 \\
			FR2 &  V3 & 0.267 & 714 & 13.0 \\
			FR2 &  V4 & 0.254 & 679 & 13.0 \\
			FR2 &  V5 & 0.338 & 904 & 13.0 \\
			\midrule
			FR3 &  V1 & 0.337 & 901 & 13.0 \\
			FR3 &  V2 & 0.290 & 776 & 13.0 \\
			FR3 &  V3 & 0.259 & 693 & 13.0 \\
			FR3 &  V4 & 0.233 & 623 & 13.0 \\
			FR3 &  V5 & 0.320 & 856 & 13.0 \\
			\bottomrule
		\end{tabular}%
	}
\end{table}
\newpage
\section{Cardiac-cycle filmstrip}\label{appendix:filmstrip}

\begin{figure}[htbp]
	\centering
	\includegraphics[width=\textwidth]{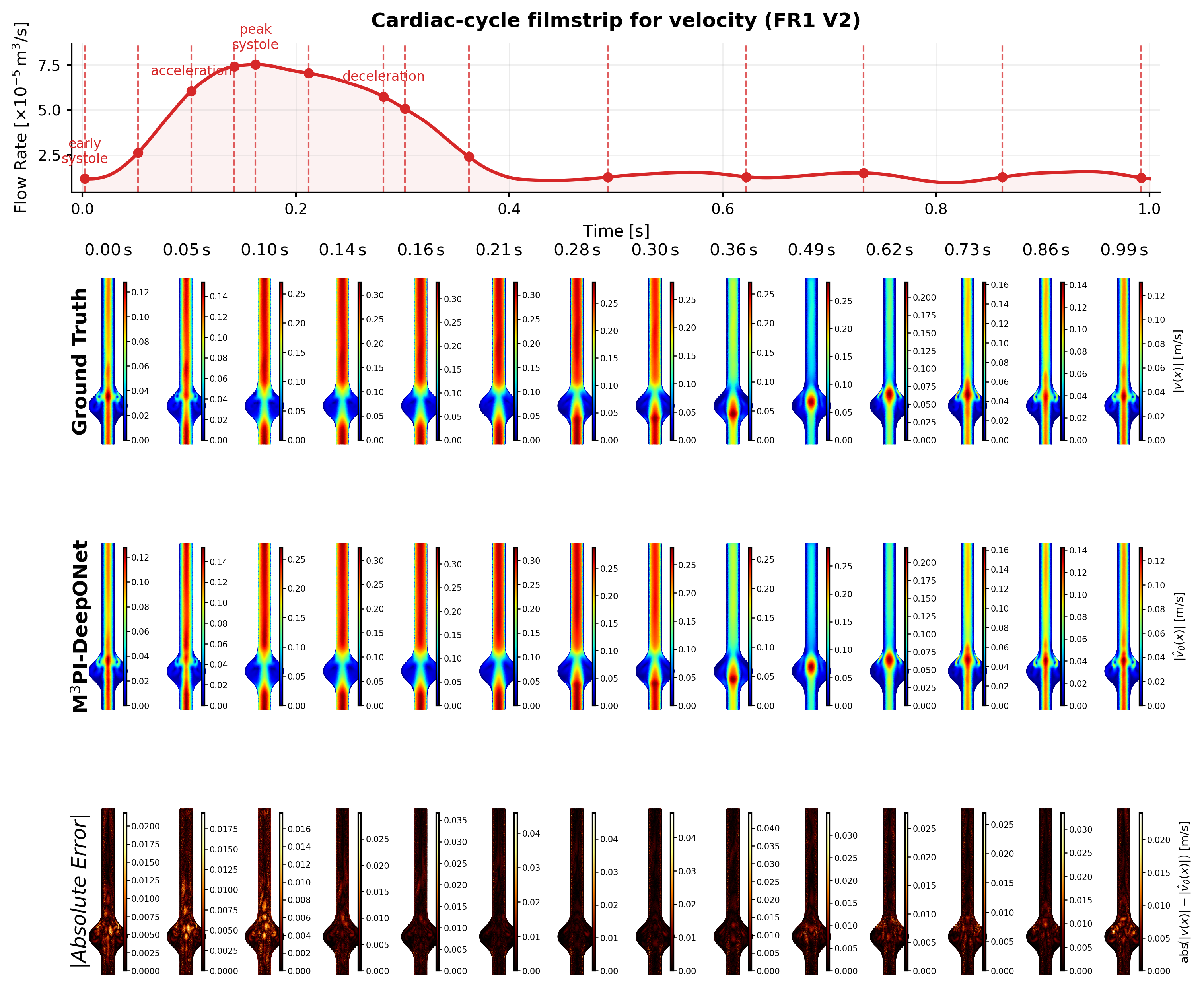}
	\caption{Cardiac-cycle filmstrip of the magnitude of velocity for the held-out test waveform FR1$ $V2. Top inset: inlet flow-rate waveform with four characteristic phases (early systole, acceleration, peak systole, deceleration) and the $14$ sampled time instants indicated by vertical markers. Each column corresponds to one of these instants. Top row: CFD ground truth; middle row: M$^{3}$PI-DeepONet prediction; bottom row: pointwise absolute error $\lvert v(\boldsymbol{x}) - \hat{v}_{\theta}(\boldsymbol{x})\rvert$. The maps are evaluated on the $x_3 = 0$ probe plane.}
	\label{fig:appendix:filmstrip_velocity}
\end{figure}

\begin{figure}[htbp]
	\centering
	\includegraphics[width=\textwidth]{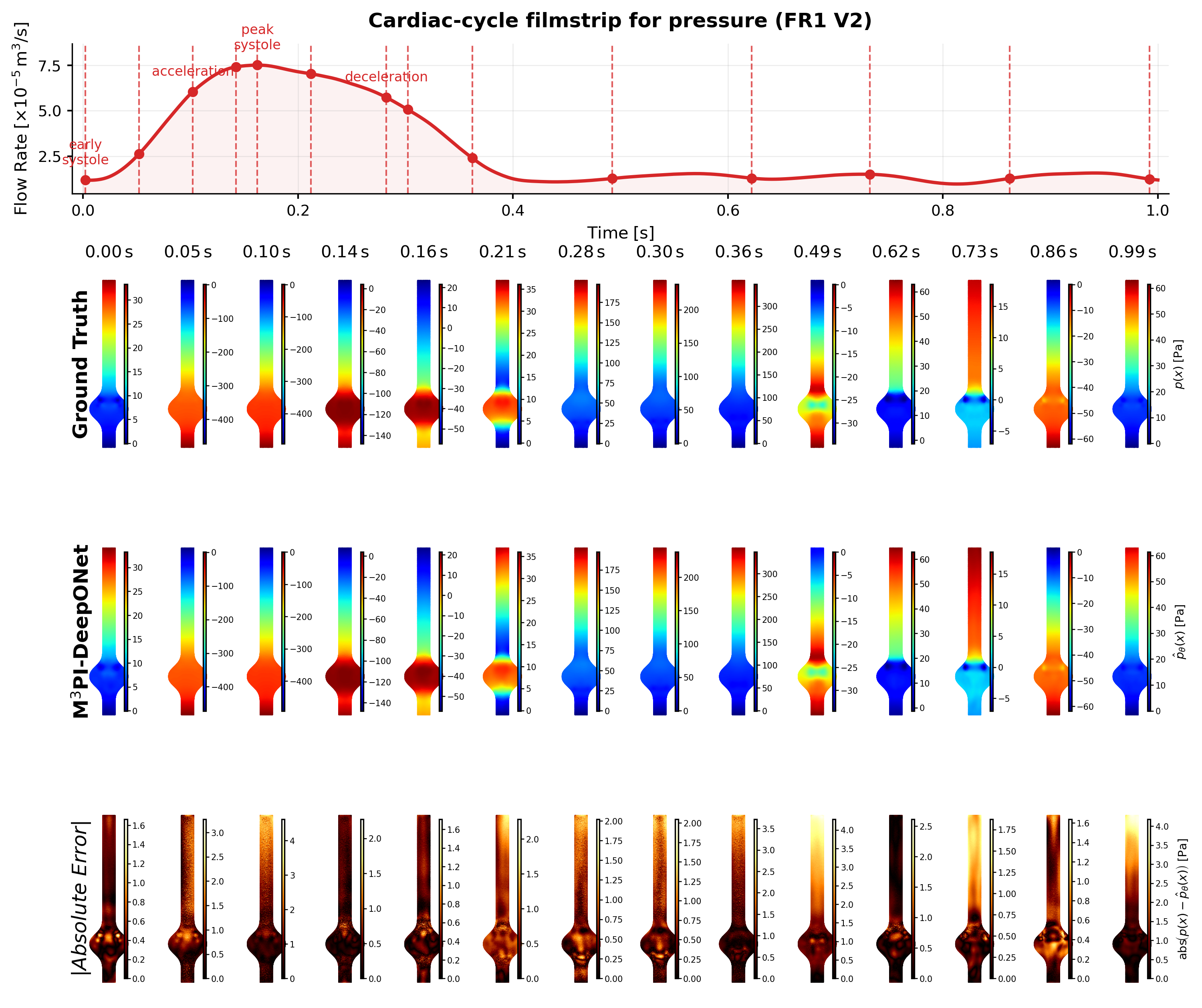}
	\caption{Corresponding cardiac-cycle filmstrip of the pressure field. Similar to Figure~\ref{fig:appendix:filmstrip_velocity}, this strip shows the evolution of the pressure field for the held-out test waveform FR1$ $V2, on the same $14$ time instants and the $x_3 = 0$ plane. Top row: CFD ground truth; middle row: M$^{3}$PI-DeepONet prediction; bottom row: pointwise absolute error $\lvert p(\boldsymbol{x}) - \hat{p}_{\theta}(\boldsymbol{x})\rvert$.}
	\label{fig:appendix:filmstrip_pressure}
\end{figure}

To complement the single-instant snapshots of Section~\ref{sect:results:final_performance}, this appendix reports filmstrips of the CFD ground truth, the M$^{3}$PI-DeepONet prediction, and the pointwise absolute error at $14$ instants of the cardiac cycle for the held-out waveform FR1$ $V2. Velocity and pressure filmstrips are shown in Figures~\ref{fig:appendix:filmstrip_velocity} and \ref{fig:appendix:filmstrip_pressure}. Equivalent filmstrips for the other two held-out waveforms (FR2$ $V3 and FR3$ $V2) display qualitatively similar behaviour and are omitted from the manuscript for conciseness.

\newpage
\bibliographystyle{elsarticle}
\bibliography{Bibliography}


\end{document}